\documentclass[lettersize,journal]{IEEEtran}
\usepackage{amsmath,amsfonts}
\usepackage{algorithmic}
\usepackage{algorithm}
\usepackage{array}
\usepackage{textcomp}
\usepackage{stfloats}
\usepackage{url}
\usepackage{verbatim}
\usepackage{graphicx}
\usepackage{float}
\usepackage{subcaption}

\usepackage{cite}
\usepackage{times}
\usepackage{soul}
\usepackage{url}
\usepackage[
    colorlinks=true,
    linkcolor=blue,
    filecolor=blue,      
    urlcolor=magenta,
    citecolor=green,]{hyperref}
\usepackage[utf8]{inputenc}
\usepackage[font=small]{caption}
\usepackage{amsmath}
\usepackage{amsthm}
\usepackage{booktabs}
\usepackage{algorithm}
\usepackage{algorithmic}
\usepackage[switch]{lineno}
\usepackage{adjustbox}
\usepackage{multirow} 
\usepackage{makecell}
\usepackage{color}
\usepackage{caption}
\usepackage{subcaption}
\usepackage{makecell}
\usepackage[table]{xcolor}
\usepackage{indentfirst}
\usepackage{amssymb}
\usepackage{caption}
\usepackage{subcaption}
\begin{document}

\title{You Only Need One Color Space:  An Efficient Network for Low-light Image Enhancement}

\author{
Qingsen Yan$^*$,
Yixu Feng$^*$,
Cheng Zhang$^*$,
Pei Wang,
Peng Wu,
Wei Dong,
Jinqiu Sun,
Yanning Zhang
% \affiliations
% School of Computer Science, Northwestern Polytechnical University
% \emails
% \{yixu-nwpu, zhangcheng233, wangpei23\}@mail.nwpu.edu.cn,\\
% \{xdwupeng, qingsenyan\}@gmail.com,
% ynzhang@nwpu.edu.cn
\thanks{Q. Yan, Y. Feng, C. Zhang, P. Wang, P. Wu, J. Sun and Y. Zhang are with the School of Computer Science, Northwestern Polytechnical University.}% <-this % stops a space
\thanks{W. Dong is with the School of Computer Science, Xi'an University of Architecture and Technology.}
}

        % <-this % stops a space
% \thanks{This paper was produced by the IEEE Publication Technology Group. They are in Piscataway, NJ.}% <-this % stops a space
% \thanks{Manuscript received April 19, 2021; revised August 16, 2021.}}

% The paper headers
% \markboth{Journal of \LaTeX\ Class Files,~Vol.~14, No.~8, August~2021}%
% {Shell \MakeLowercase{\textit{et al.}}: A Sample Article Using IEEEtran.cls for IEEE Journals}

% \IEEEpubid{0000--0000/00\$00.00~\copyright~2021 IEEE}
% Remember, if you use this you must call \IEEEpubidadjcol in the second
% column for its text to clear the IEEEpubid mark.

\maketitle
\renewcommand{\thefootnote}{}
\footnotetext{$^*$These authors contributed equally to this work. Corresponding author: Yanning Zhang} 
\begin{abstract}
Low-Light Image Enhancement (LLIE) task tends to restore the details and visual information from corrupted low-light images.
    Most existing methods learn the mapping function between low/normal-light images by deep neural networks on sRGB and HSV color space.
    %Nevertheless, enhancement involves amplifying image signals, and employing these color spaces can render the enhancement process sensitive and unstable, resulting in the emergence of color artifacts and brightness artifacts.
    % 
    % Nevertheless, enhancement involves amplifying image signals, and applying these color spaces to low-light images with a low signal-to-noise ratio can introduce sensitivity and instability into the enhancement process. Consequently, this results in the presence of color artifacts and brightness artifacts in the enhanced images.
    % %% sen
    However, these methods involve sensitivity and instability in the enhancement process, which often generate obvious color and brightness artifacts.
    % %%
    %However, sRGB space has a strong coupling of the color axis (\emph{i.e.}, Red, Green, Blue), which makes it hard to recover the content from low-light images. Although retinex-based methods try to decouple illumination and reflectance from sRGB space, it is challenging in low-light regions due to noise and color deviation. 
    To alleviate this problem, we propose a novel trainable color space, named Horizontal/Vertical-Intensity (\textbf{HVI}), which not only decouples brightness and color from RGB channels to mitigate the instability during enhancement, but also adapts to low-light images in different illumination ranges due to the trainable parameters.
    %To alleviate this problem, we propose a trainable color space, named Horizontal/Vertical-Intensity (HVI), to reduce the difficulty of feature extraction from sRGB color space. The proposed HVI consists of an intensity axis to represent the illumination intensity, and H, V represent a trainable color plane.
    Furthermore, we design a novel Color and Intensity Decoupling Network (\textbf{CIDNet}) with two branches dedicated to processing the decoupled image brightness and color in the HVI space.
    In addition, we introduce the Lighten Cross-Attention (LCA) module to facilitate interaction between image structure and content information in both branches, while also suppressing noise in low-light images.
    %brightness and color in to effectively use HVI space to process and enhance low-light images, consisting of Lighten Cross-Attention (LCA) module and two-way UNet. Specifically, the LCA block uses bidirectional cross-attention to learn the strong relationship between low intensity and color.
    %In addition, we design value-enhancement and denoise color layers in LCA block to recover color and remove noise in the low-light regions, respectively. Finally, we convert the image in HVI space to the sRGB space to obtain a normal-light image.  
    We conduct 22 quantitative and qualitative experiments to show that the proposed CIDNet outperforms the state-of-the-art methods on 11 datasets. The code is available at \href{https://github.com/Fediory/HVI-CIDNet}{https://github.com/Fediory/HVI-CIDNet}.
\end{abstract}

\begin{IEEEkeywords}
Low-Light Image Enhancement, HVI Color Space, Transformer, Supervised Learning.
\end{IEEEkeywords}

\section{Introduction}
\label{section:Introduction}

Under low-light conditions, the sensor captures weak light signals with severe noise, resulting in poor visual quality for low-light images.
Obtaining high-quality images from degraded images often necessitates Low-Light Image Enhancement (LLIE), intending to improve the brightness while simultaneously reducing the impact of noise and color bias \cite{2022LLE}.

\begin{figure}
    \centering
    \includegraphics[width=1\linewidth]{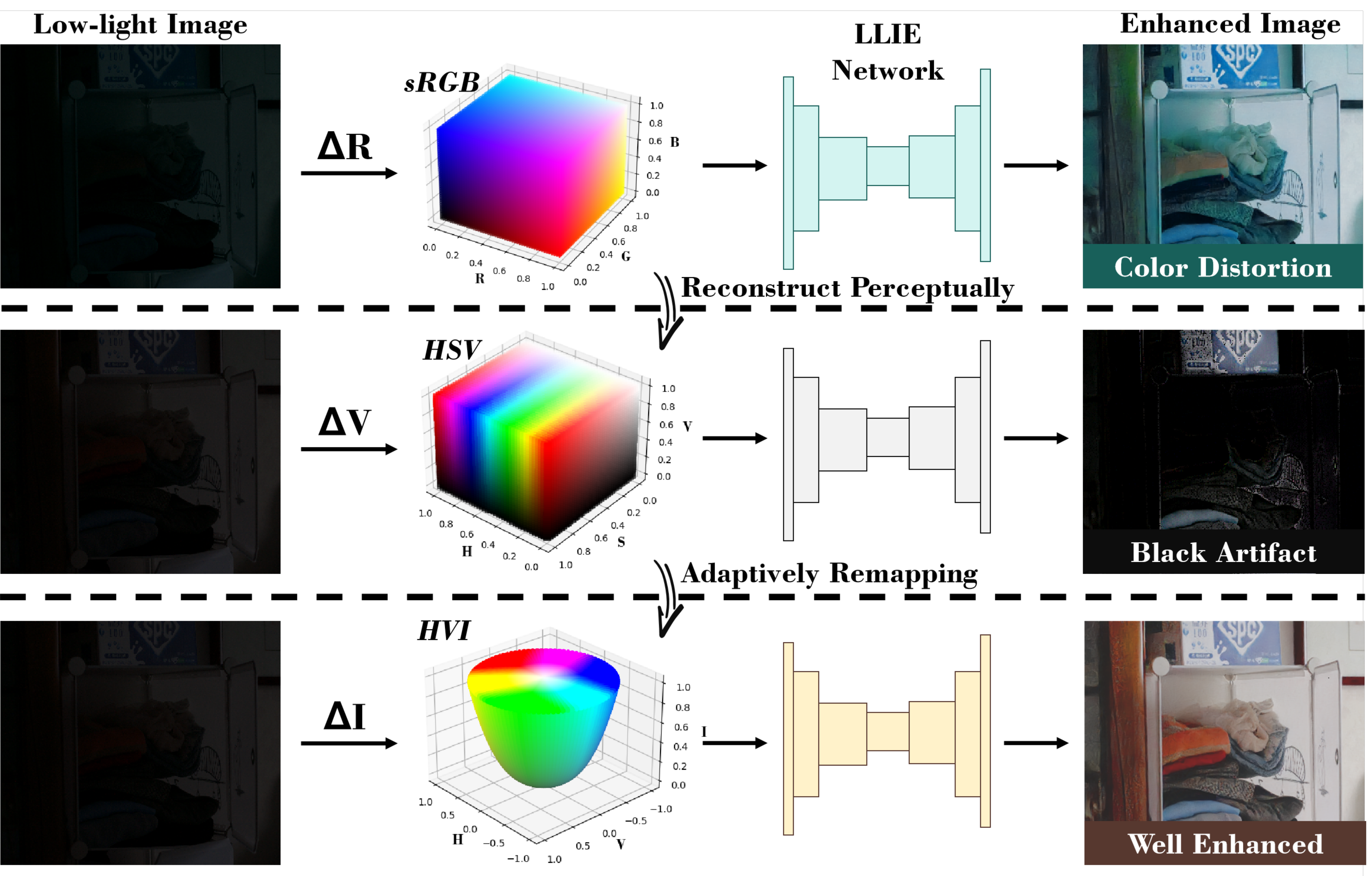}
    \caption{The sensitivity comparison of different color spaces in low-light enhancement. The notations $\Delta$R, $\Delta$V, and $\Delta$I represent a tiny variation in the axis of Red (sRGB), Value (HSV), and Intensity (HVI), respectively.  After enhancement processing, noticeable color artifacts can be observed in the sRGB and HSV space results. }
    \label{fig:1}
\end{figure}

The majority of existing LLIE approaches \cite{KinD, EnGAN,Zero-DCE} focus on finding an appropriate image brightness, and commonly employ deep neural networks to learn the mapping relationship between low-light images and normal-light images within the sRGB space.
However, image brightness and color exhibit a strong interdependence with the three channels in sRGB. 
A slight disturbance in the color space will cause an obvious variation in both the brightness and color of the generated image.
% However, information such as image brightness and color exhibits a strong interdependence with the three channels in sRGB. 
% Noise in any channel significantly impacts both the brightness and color of the image. 
As illustrated in Fig. \ref{fig:1}, introducing noise to one dimension (red channel) in the sRGB space dramatically changes the color of the enhanced image. 
This suggests a mismatch between sRGB space and the low-light enhancement processing, resulting in instability in both brightness and color in the enhanced results.
% The reason can be attributed to the inherent instability that many enhancement methods \cite{LLFlow,EnGAN} require more parameters and complex structures to learn this enhancement mapping.
The inherent instability leads to the existing enhancement methods \cite{LLFlow,EnGAN} requiring more parameters and complex network architecture to learn this enhancement mapping.
It is also why numerous low-light enhancement methods \cite{RetinexNet,PairLIE} need to incorporate additional brightness and color losses during training.

While the HSV color space \cite{Bread} enables the separation of brightness and color of the image from sRGB channels, the discontinuous property of hue axis (see Sec. \ref{sec:HVI}-B) and its intricate mapping relationships with sRGB space makes it challenging to handle complex and varying lighting conditions. As shown in Fig. \ref{fig:1}, 
% enhanced images using the HSV space often result in black artifacts due to even minor noise. 
enhanced images with the HSV space often have obvious black artifacts due to extremely low light environments. 
We consider that the color space (\emph{e.g.}, sRGB, HSV) has a huge impact on the image enhancement effect.

To address the aforementioned issue between the low-light image enhancement task and existing color spaces, we introduce a new color space named \textbf{H}orizontal/\textbf{V}ertical-\textbf{I}ntensity (\textbf{HVI}), designed specifically to cater to the needs of low-light enhancement tasks. 
The proposed HVI color space not only decouples brightness and color information but also incorporates three trainable representation parameters and a trainable function, allowing it to adapt to the brightening scale and color variations of different low-light images.
Building upon the HVI color space, to fully leverage the decoupled information, we propose a new LLIE method, named \textbf{C}olor and \textbf{I}ntensity \textbf{D}ecoupling Network (\textbf{CID}Net).
CIDNet consists of HV-branch and intensity-branch, which makes full use of decoupled information to generate high-quality results. After applying the HVI transformation to the image, it is separately fed into the HV-branch to extract color information, and the intensity-branch to establish the photometric mapping function under different lighting conditions.
Additionally, to enhance the interaction between the structures of images contained in the brightness and color branches, we propose the bidirectional Lighten Cross-Attention (LCA) module to learn the complementary information of HV-branch and intensity-branch.
Furthermore, we conduct experiments and ablation studies on multiple datasets to validate our approach. The experimental results demonstrate that CIDNet effectively enhances the brightness of low-light images while preserving their natural colors, which validates the compatibility of the proposed HVI color space with low-light image enhancement tasks. Note that the proposed method exhibits relatively small parameters (1.88M) and computational loads (7.57G), achieving a good balance between effectiveness and efficiency on edge devices.

Our contributions can be summarized as follows:
\begin{enumerate}
    \item We introduce a novel HVI color space with trainable parameters, which not only decouples the image brightness and color, but also adapts to various image illumination scales.
    \item Based on the HVI color space, we propose a dual-branch network, CIDNet, to concurrently process the brightness and color of low-light images.
    \item We design a LCA module to facilitate interaction between the HV-branch and intensity-branch, allowing the scene information in each branch to complement and improve the visual effects of the enhanced image. 
    \item We conduct 22 quantitative and qualitative experiments that our CIDNet outperforms all types of SOTA methods on 10 different metrics across 11 datasets. 
\end{enumerate}

\section{Related Work}

\subsection{Low-Light Image Enhancement}
\textbf{Traditional and Plain LLIE Methods.} Plain methods usually enhanced image by histogram equalization \cite{2007HE} and Gama Corrction \cite{2013GC}. Traditional method \cite{2016R} commonly depends on the application of the Retinex theory, which decomposes the lights into illumination and reflections. For example, Guo \textit{et al.} \cite{LIME} refine the initial illumination map to optimize lighting details by imposing a structure prior. Regrettably, existing methodologies are inadequate in effectively eliminating noise artifacts and producing accurate color mappings, rendering them incapable of achieving the desired level of precision and finesse in the LLIE tasks.
\par
\textbf{Deep Learning Methods.} Deep learning-based approaches \cite{LLNet,KinD,EnGAN,RetinexFormer} has been widely used in LLIE task. 
Existing methods propose distinct solutions to address the issues of image color shift and noise stabilization. 
For instance, RetinexNet \cite{RetinexNet} enhances images by decoupling illumination and reflectance based on Retinex theory. However, it has unsatisfied results with several color shifts.
% RUAS \cite{RUAS} unrolled with architecture search to construct Lighten yet effective LLIE network. 
SNR-Aware \cite{SNR-Aware} presents a collectively exploiting Signal-to-Noise-Ratio-aware transformers to dynamically enhance pixels with spatial-varying operations, which could reduce the color bias and noise. 
Bread \cite{Bread} decouples the entanglement of noise and color
distortion by using YCbCr color space. Furthermore, they designed a color adaption network to tackle the color distortion issue left in light-enhanced images. 
Still, SNR-Aware and Bread show poor generalization ability. They are not only inaccurately controlled in terms of brightness in some of the datasets, but also biased in terms of color with pure black area.
% Still, all of these methods are recovered on sRGB space, which is not only inaccurately controlled in terms of brightness, but also biased in terms of color.

\textbf{Diffusion Model-based Methods.} With the advancement of Denoising Diffusion Probabilistic Models (DDPMs) \cite{Ho2020Denosing}, diffusion-based generative models have achieved remarkable results in the LLIE task. It has indeed shown the capability to generate more accurate and appropriate images in pure black spaces devoid of information and under low light conditions with significant noise. However, they still exhibit issues such as local overexposure or color shift. Recent LLIE diffusion methods have attempted to address these challenges by incorporating global supervised brightness correction or employing local color correctors \cite{zhou2023pyramid,wu2023reco,hou2024global}.
PyDiff \cite{zhou2023pyramid} employs a Global Feature Modulation to correct the pixel noise and color bias globally.
Diff-Retinex \cite{yi2023diff} rethink the retinex theory with a diffusion-based model in the LLIE task, which decomposed an image to illumination and reflectance colors to reduce color bias and enhance brightness separately. However, the aforementioned diffusion models suffer from long training and inference times, lack of Lighten efficiency, and inability to fully decouple brightness and color information.
% DiffPIR \cite{zhu2023denoising} innovatively integrates traditional plug-and-play image restoration methods with diffusion model, achieved the state-of-the-art reconstruction fidelity and perceptual quality, while maintaining a low count of neural function evaluations.
% \textbf{(2)} Zero-shot \cite{choi2021ilvr,Chung2021Come,Lugmayr2022repaint}, where pre-trained generative models are considered repositories of structures and textures constructed from extensive real datasets, thus leveraging pre-trained diffusion models to acquire structural and textural priors for image restoration.

\subsection{Color Space}

\label{section:CS}

\textbf{RGB.} 
% Any additive color space based on RGB color model belongs to the RGB color space. 
Currently, the most commonly used is the standard-RGB (sRGB) color space. For the same principle as visual recognition by the human eye, sRGB is widely used in digital imaging devices \cite{POYNTON2003187}. Nevertheless, image brightness and color exhibit a strong interdependence with the three channels in sRGB.
A slight disturbance in the color space will cause an obvious variation in both the brightness and color of the generated image. Thus, sRGB is not the optimal color space for enhancement.

\textbf{HSV and YCbCr.} Hue, Saturation and Value (HSV) color space represents points in an RGB color model with a cylindrical coordinate system \cite{Foley1982FundamentalsOI}. Indeed, it does decouple brightness and color of the image from RGB channels. However, the inherent hue axis color discontinuity and non-mono-mapped pure black planes pose significant challenges when attempting to enhance the image in HSV color space, resulting in the emergence of obvious artifacts. To circumvent issues related to HSV, some methods \cite{Bread,brateanu2024lytnet} also transform sRGB images to the YCbCr color space for processing, which has an illumination axis (Y) and reflect-color-plain (CbCr). Although it solved the hue dimension discontinuity problem of HSV, the multi-mapping of pure black planes still exists.

% \textbf{YCbCr and CIELAB.} They are all interconvertible with RGB space and also solve the hue dimension discontinuity problem of HSV \cite{Foley1982FundamentalsOI}. Still, this does not solve the problem of multi-mapping of pure black planes, which is a reason why Bread method \cite{Bread} have a poor generation ability. 

\section{HVI Color Space}
\label{sec:HVI}
To address the misalignment between LLIE and existing color spaces, we innovatively introduce a trainable color space in the field of LLIE, named Horizontal/Vertical-Intensity (HVI) color space. HVI color space consists of three trainable parameters and a custom training function that can adapt to the photosensitive characteristics and color sensitivities of low-light images. In this section, we will present the description of the mono-mapping image transformation from sRGB space to HVI color space, as well as the details of HVI color space. \par

%To sort out the aforementioned color space challenges, we first present a trainable Horizontal/Vertical-Intensity (HVI) color space. It consists of three trainable parameters and a custom training function that can adapt to the photosensitive characteristics and color sensitivities of the dataset. Specifically, our focus lies on developing a mono-mapping transform that enables the conversion between sRGB and HVI. \par
\subsection{Intensity Map}
In the task of LLIE, one crucial aspect is accurately estimating the illumination intensity map of the scene. Following retinex-based LLIE methods \cite{LIME, RetinexNet, PairLIE}, we represent the intensity map of an image with the maximum value among the RGB channels.
% , as all channels in an image share the same illumination. 
According to the definition, we can calculate the intensity map of an image $\mathbf{I}_{max} \in \mathbb{R}^{\mathrm{H\times W\times 3}} $ as follows:
\begin{equation}
\mathbf{I}_{max} =\max_{\mathbf{c}\in \{R,G,B\}} (\mathbf{I_{c}}),  
\label{eq:1}
\end{equation}
Next, we will introduce how to utilize the intensity map to guide the generation of the corresponding HV color map.
%Any single sRGB image $\mathbf{I} \in \mathbb{R}^{\mathrm{H\times W\times 3}} $ can be decomposed into three image $\mathbf{I_{c}} \in \mathbb{R}^{\mathrm{H\times W} }$ where $\mathbf{c}\in \{ R,G,B\}$ . There we denote the pixel point light intensity by 
%\begin{equation}
%\mathbf{I}_{max} =\max_{\mathbf{c}\in \{R,G,B\}} (\mathbf{I_{c}}),  
%\label{eq:1}
%\end{equation}
%which represent intensity map. 

\subsection{HV Transformation}
\begin{figure}
    \centering
    \includegraphics[width=1\linewidth]{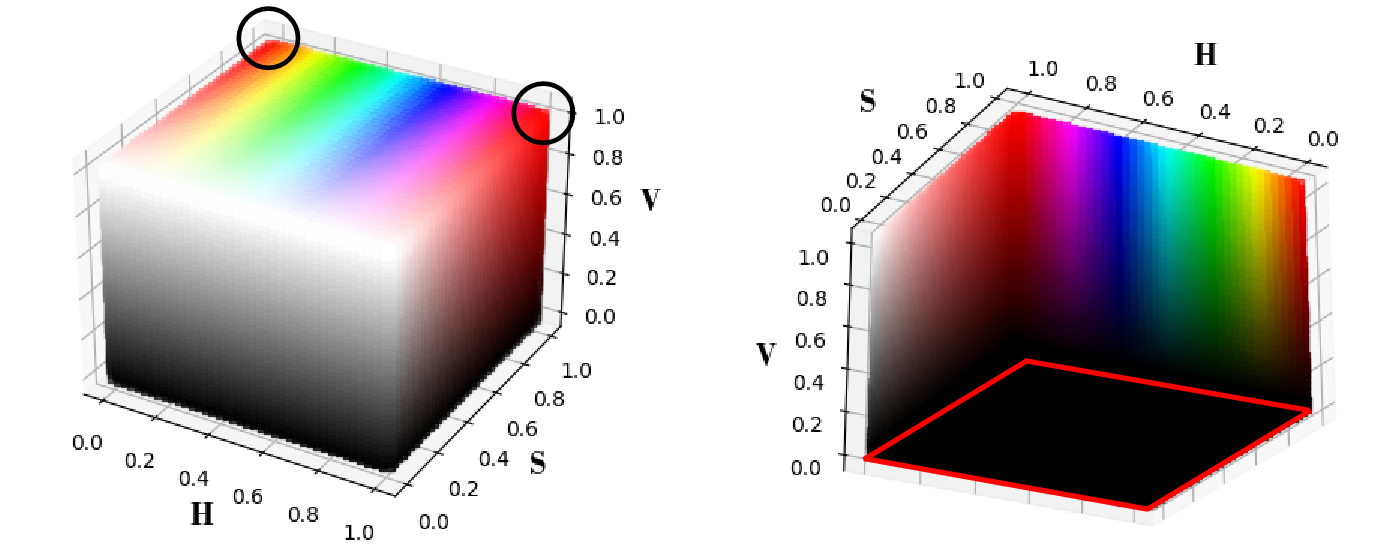}
    \caption{HSV color space visualization. The two black circles in the left image indicate the discontinuous color positions along the hue axis, while the red box in the right image displays a pure black plane with Value$=0$.}
    \label{fig:HSV}
\end{figure}

% 加入RGB部分，由retinex中reflect引入HSV，后面写出HSV的问题
To effectively address color deviations, previous LLIE methods typically separate an sRGB image into a reflectance map \cite{RetinexNet, KinD} based on the retinex theory. These approaches often require a large computational network for fitting, which need to use a combination of hue and saturation to simulate the reflectance map. 
% The transformation of RGB to saturation ($\mathbf{S}$) is defined as follows
% \begin{equation}
% \mathbf{S}  = \begin{cases}
% 0, &  \mathbf{I}_{max}   = \mathbf{0} \\
% \frac{\Delta}{\mathbf{I}_{max} }, &  \mathbf{I}_{max}  \ne \mathbf{0} \\
% \end{cases} \\
% ,
% \end{equation}
% where $\Delta=\mathbf{I}_{max}-min(\mathbf{I}_c)$ and the hue axis ($\mathbf{H}$) is formulated as
% \begin{equation}
% \mathbf{H}= \begin{cases}
% 0, &\text{if } \mathbf{S} = 0 \\
% \frac{\mathbf{I_{G}} - \mathbf{I_{B}}}{\Delta} \mod{6},  &\text{if } \mathbf{I}_{max}  = \mathbf{I_{R}} \\
% 2+\frac{\mathbf{I_{B}} - \mathbf{I_{R}}}{\Delta},  &\text{if } \mathbf{I}_{max}   = \mathbf{I_{G}} \\
% 4+\frac{\mathbf{I_{R}} - \mathbf{I_{G}}}{\Delta},  &\text{if } \mathbf{I}_{max}   = \mathbf{I_{B}} 
% \end{cases}
% ,
% \end{equation}
% where $\mathbf{I_{R}},\mathbf{I_{G}},\mathbf{I_{B}}\in \mathbb{R}^{\mathrm{H\times W}} $. 
However, the HSV color space is discontinued in the hue axis (see the black circles in Fig. \ref{fig:HSV}) and has a non-bijection pure black plane (see the red rectangle in Fig. \ref{fig:HSV}), which disrupts the one-to-one mapping. 
% Thus, the HSV is discontinuous on the hue axis. 
For discontinuity of the hue axis, based on HSV color transform formula, a red color in sRGB $(R,G,B)=(1,0,0)$ corresponds to $(\textcolor{red}{0},1,1)$ and $(\textcolor{red}{1},1,1)$ in HSV color space.
% an pixel with $(\textcolor{red}{H},S,V)=(\textcolor{red}{0},1,1)$ is the same as $(\textcolor{red}{H},S,V)=(\textcolor{red}{1},1,1)$ in Fig. \ref{fig:HSV}), which represents the red color in sRGB $(R,G,B)=(1,0,0)$.
% Furthermore, as shown in Fig. \ref{fig:HSV}, any HSV pixel with Value$=0$ denote the black color, which can be described in sRGB color space by $(R,G,B)=(0,0,0)$. 
For the pure black plane, as shown in Fig. \ref{fig:HSV}, any HSV pixel with Value$=0$ denote the black color, corresponding to $(R,G,B)=(0,0,0)$ in sRGB color space.
Thus, the one-to-many mapping is the source reason of why the existing methods on HSV color space generate the obvious artifacts in the dark regions.
% causes the emergence of highly pronounced artifacts in the final image.

To address the suboptimum problems caused by one-to-many mapping (HSV color space), we design a trainable Horizonal/Vertical (HV) color map as a plane to quantify a color-reflectance map, which is a one-to-one mapping to sRGB color space. The HVI color space consists of three trainable parameters $k$, $\gamma_G$ and $\gamma_B$, and a custom training function $T(x)$.

\textbf{Parameter $k$.} 
Considering that the dark regions of low-light images have small values, which is hard to distinguish the color and causes information loss.
Based on the proposed HVI color space, we customize a parameter $k$ that allows networks to adjust the color point density of the low-intensity color plane, which quantifies a Color-Density-$k$ ($\mathbf{C}_k$) as
% Inspired by the weakness of the tiny value of low-lights in sRGB, we customize a parameter $k$ that allows networks to adjust the color point density of the low-intensity color plane, which quantifies a Color-Density-$k$ ($\mathbf{C}_k$) as
\begin{equation}
\mathbf{C}_k=\sqrt[k]{\sin (\frac{ \pi \mathbf{I}_{max} }{2} )+\mathcal{\varepsilon} },
\label{eq:2}
\end{equation}
where $k\in \mathbb{Q^+}$ and we set $\mathcal{\varepsilon}=1\times10^{-8}$.

\textbf{Hue bias parameter $\gamma_G$ and $\gamma_B$.} 
Since different cameras have different sensitivity on RGB channel, it will cause color shift (\emph{i.e.} green color) on the low-light scenes.
To alleviate the data diversity caused by color shift, we learn an adaptive linear Color-Perceptual map $\mathbf{P}_\gamma$ on hue value.
\begin{equation}
\mathbf{P}_\gamma= \begin{cases}
3\gamma_G\mathbf{H},  &\text{if } 0\le \mathbf{H} <\frac{1}{3}\\
3(\gamma_B-\gamma_G)(\mathbf{H}-\frac{1}{3})+\gamma_G,  &\text{if } \frac{1}{3}\le\mathbf{H} <\frac{2}{3} \\
3(1-\gamma_B)(\mathbf{H}-1)+1,  &\text{if } \frac{2}{3}\le \mathbf{H} \le1
\end{cases}
,
\label{eq:5}
\end{equation}
where $\gamma_G,\gamma_B\in(0,1)$, $\mathbf{H}\in\left[ 0,1\right]$ denotes the hue value. 

\textbf{Customizable or trainable function $T(x)$.} 
% Previous 
% The saturation of results generated by previous methods are 
To improve the saturation the generated results, we design a Function-Density-$T$ based on the $\mathbf{P}_\gamma$ to adaptively adjust the saturation.
% To enhance the functionality of our color space,  
We utilize a Function-Density-$T$ as
\begin{equation}
    \mathbf{D}_T = T(\mathbf{P}_\gamma),
    \label{eq:6}
\end{equation}
% where $T$ is an elemental function as $T(x)$ ($x$ represents an element in a matrix), which is defined only $x\in\left[ 0,1\right]$ and satisfies $T(0)=T(1)$ and $T(x)\ge0$. 
where $T(\cdot)$ satisfies $T(0)=T(1)$ and $T(\mathbf{P}_\gamma)\ge0$. 
Finally, we formalize the horizontal ($\mathbf{\hat{H}}$) and vertical ($\mathbf{\hat{V}}$) plane as
\begin{equation}
\begin{split}
    \mathbf{\hat{H}} = \mathbf{C}_k \odot  \mathbf{S}   \odot \mathbf{D}_T\odot h,\\
    \mathbf{\hat{V}} = \mathbf{C}_k \odot  \mathbf{S}   \odot \mathbf{D}_T\odot v,
\end{split}
\label{eq:7}
\end{equation}
where $\odot$ denotes the element-wise multiplication. Note that we orthogonalize our color map by setting an intermediate variable 
$h= \cos (2\pi \mathbf{P}_\gamma)$ and $v= \sin (2\pi \mathbf{P}_\gamma)$ to be bijective.

% \subsection{Trainable HVI Map}
% \label{sec:3-3}
% To ensure the integrity of image information and prevent data loss, we concatenate $\mathbf{\hat{H}}$, $\mathbf{\hat{V}}$, $\mathbf{I}_{max}$ (Eq. \ref{eq:1} and \ref{eq:7}) to our $\mathbf{I_{HVI}} \in \mathbb{R}^{\mathrm{H\times W\times 3}} $ map. Simultaneously, the HVI map enables a mono-correspondence between each color point in the sRGB and HVI color spaces to ensure the reversibility of the transformation. 
% % This map can be recognized by computer perceptually. 
% % Not only solve it the problem of multi-mapping of pure black planes and intermittent color axis, but moreover, it can adapt itself to a variety of tasks by way of machine learning for more functionality.
% The HVI color spaces not only resolves the challenges of multi-mapping but also possesses the ability to adapt to a diverse array of tasks through machine learning. 

% % This integration of machine learning not only enhances functionality but also offers the potential for continual improvement and adaptation to various scenarios within the domain.

\subsection{Perceptual-invert HVI Transformation}
\label{sec:3-4}
To convert HVI back to the HSV color space, we design a Perceptual-invert HVI Transformation (PHVIT), which is a surjective mapping while allowing for the independent adjustment of the image’s saturation and brightness.

To transform injectively, the PHVIT sets $\hat{h}$ and $\hat{v}$ as an intermediate variable as 
\begin{equation}
\hat{h}=\frac{ \mathbf{\hat{I}_{H}}}{\mathbf{D}_T\mathbf{C}_k+\mathcal{\varepsilon}},
\hat{v}=\frac{ \mathbf{\hat{I}_{V}}}{\mathbf{D}_T\mathbf{C}_k+\mathcal{\varepsilon}},
\end{equation}
where $\mathcal{\varepsilon}=1\times10^{-8}$. 
% To simplify the inverse steps, we have chosen to convert $\hat{h}$ and $\hat{v}$ to HSV color space. 
Then, we convert $\hat{h}$ and $\hat{v}$ to HSV color space.
The hue map can be formulated by 
\begin{equation}
    H = F_\gamma(\arctan(\frac{\hat{v}}{\hat{h}})\mod 1),
\end{equation}
where $F_\gamma$ is a 
% $\mathbf{P}_\gamma-\mathbf{H}$
inverse linear function as 
\begin{equation}
    F_\gamma(\mathbf{X}) = \begin{cases}
        \frac{\mathbf{X}}{3\gamma_G}, &\text{if } 0 \le \mathbf{X} < \gamma_G \\
        \frac{\mathbf{X}-\gamma_G}{3(\gamma_B-\gamma_G)} + \frac{1}{3}, &\gamma_G \le \mathbf{X} < \gamma_B\\
        \frac{\mathbf{X}-1}{3(1-\gamma_B)} + 1 ,& \gamma_B \le \mathbf{X} \le 1
    \end{cases}
    ,
\end{equation}
where $\gamma_G,\gamma_B$ are mentioned in Eq. \ref{eq:5}. The saturation and value map can be perceptually estimated as 
\begin{equation}
\begin{split}
    S &= \alpha_{S}\sqrt{\hat{h}^{2}+\hat{v}^{2}},\\
    V &= \alpha_{I}\hat{\mathbf{I}}_{\mathbf{I}},
\end{split}
\label{eq:10}
\end{equation}
where $\alpha_{s},\alpha_{i}$ are the customizing linear parameters to change the image color saturation and brightness. 
% Intensity map $\widetilde{\mathbf{I}}_{\mathbf{\hat{I}}}$ can be measured as Value, $\mathbf{\hat{C}_h}$ as hue and $\mathbf{\hat{C}_s}$ as Saturation in HSV. 
% The concatenation of $H$ (Hue), $S$ (Saturation) and $V$ (Value) can be estimated as an HSV image, which can be converted to an sRGB image by the formula mentioned in \cite{Foley1982FundamentalsOI}.
Finally, we will obtain the sRGB image with HSV image \cite{Foley1982FundamentalsOI}.

\begin{figure*}
    \centering
    \includegraphics[width=1\linewidth]{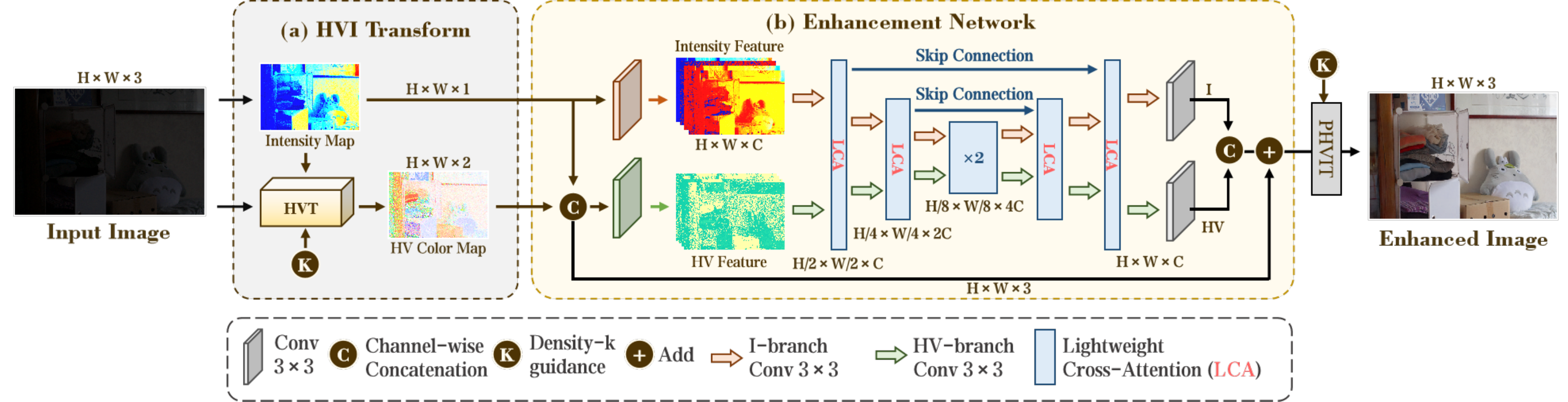}
    \caption{The overview of the proposed CIDNet. \textbf{(a)} HVI Color Transformation (HVIT) takes an sRGB image as input and generates HV color map and intensity map as outputs. \textbf{(b)} Enhancement Network performs the main processing, utilizing a dual-branch UNet architecture, which contains six Lighten Cross-Attention (\textcolor[RGB]{255,95,95}{LCA}) blocks. Lastly, we apply Perceptual-inverse HVI Transform (PHVIT) to take a light-up HVI map as input and transform it into an sRGB-enhanced image.}
    \label{fig:CIDNet-pipeline}
\end{figure*}

\section{CIDNet}
% 需要设计一个更有效果的，每条支路分别处理的网络，与HVI空间相匹配
% The goal of LLIE tasks is to map brightness to appropriate ranges while reducing noise and color shifts. In the HVI color space, brightness is estimated as an Intensity map, while denoising and color correction are performed in the HV map.
% In order to better apply the proposed HVI color space to LLIE tasks, in this section, we will introduce the pipeline of Color and Intensity Decoupling Network (CIDNet) and its structure, which is a Lighten and dual-branch model.

%基于提出的HVI空间，我们提出了一个与之相适配的dual-branch 的LLIE的网络，命名为Color and Intensity Decoupling Network（CIDNet）.CIDNet的核心思想是在HVI空间中分开处理亮度估计和噪声与色差抑制，进而提升整体的增强性能。同时，我们也在网络中引入了轻量化的交叉注意力模块用于HV-branch和I-branch之间的交互guidance. 在这一部分，我们详细介绍CIDNet的流程与CIDNet的具体结构。
Based on the proposed HVI space, we introduce a novel dual-branch LLIE network, named the Color and Intensity Decoupling Network (CIDNet) to separately address HV-plain and I-axis information in the HVI space.
The CIDNet employs HV-branch to suppress the noise and chromaticity in the dark regions and utilizes I-branch to estimate the luminance of the whole images.
Furthermore, we design an Lighten Cross-Attention (LCA) module to facilitate interaction guidance between the HV-branch and I-branch. In this section, we provide a detailed description of the architecture of CIDNet.
% . The core concept of CIDNet is to separately address luminance estimation and noise and chromaticity suppression in the HVI space, thereby enhancing the overall enhancement performance. Additionally, we design Lighten Cross-Attention (LCA) modules in the network to facilitate interaction guidance between the HV-branch and I-branch. In this section, we provide a detailed exposition of the workflow and specific architecture of CIDNet.

\begin{figure}
    \centering
    \includegraphics[width=1\linewidth]{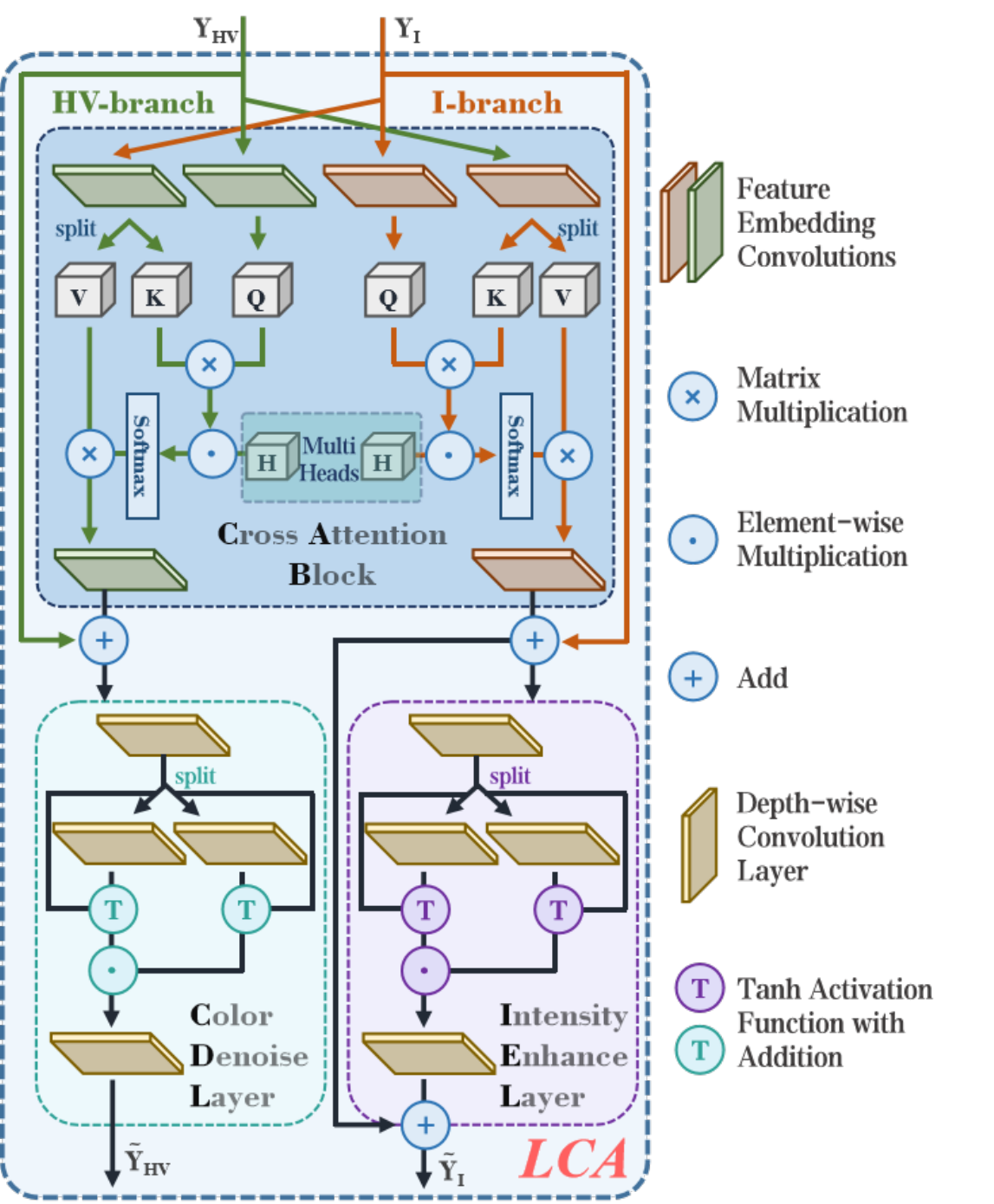}
    \caption{The dual-branch Lighten Cross-Attention (\textcolor[RGB]{255,95,95}{LCA}) block (\emph{i.e.}, I-branch and HV-branch). The LCA incorporates a Cross Attention Block (CAB), an Intensity Enhance Layer (IEL), and a Color Denoise Layer (CDL). The feature embedding convolution layers contains a $1\times1$ depth-wise convolution and a $3 \times 3$ group convolution.}
    \label{fig:CIDNet-structure}
\end{figure}

\subsection{Pipeline}

%我们方法的整体流程可分为三个串联的主要步骤，分别是图像的HVI变换,CIDNet的并行处理，HVI逆变换，如fig2所示。
As shown in Fig. \ref{fig:CIDNet-pipeline}, the overall framework can be divided into three consecutive main steps, \emph{i.e.}, HVI transformation, enhancement network, and perceptual-inverse HVI transformation.

%现有绝大多数图像均以sRGB空间存储，所以我们的输入为RGB图像$I$。如III部分所述，HVI变换将RGB图像分成两种成分，分别是包含场景照度信息的Intensity map和包含场景颜色和结构等信息的HV color map.详细来说，我们首先使用公式（1）来计算intensity map，即$\mathbf{I_{I}}=\mathbf{I}_{max}$. 然后将得到的intensity map和原始图像I 利用公式(7)来计算HV color map. As illustrated in Fig2(a)，我们在HVI空间中引入的可训练的density-k,用于调节变换过程中的颜色density和low-intensity color plane。为了实现residual learning，我们concat intensity map和HV color map 得到了保存图像原始HVI空间信息的HVI map。
% The vast majority of existing images are stored in the sRGB color space, hence our input is an RGB image $\mathbf{I}$. 

As described in Sec. \ref{sec:HVI}, the HVI transformation decomposes the sRGB image into two components: an intensity map containing scene illuminance information and an HV color map containing scene color and structure information. Specifically, we first calculate the intensity map using Eq. \ref{eq:1}, which is $\mathbf{I_{I}}=\mathbf{I}_{max}$. Subsequently, we utilize the intensity map and the original image to generate HV color map using Eq. \ref{eq:7}.
Furthermore, a trainable density-$k$ is employed to adjust the color point density of the low-intensity color plane, as shown in Fig. \ref{fig:CIDNet-pipeline}(a).
% As depicted in Fig. \ref{fig:CIDNet-pipeline}(a), a trainable density-$k$ introduced in the HVI space is employed to adjust the color density and low-intensity color plane during the transformation process. 
% To facilitate residual learning, we the concatenation of the intensity map and HV color map results in the HVI map that preserves the original image’s information in the HVI space.

%CIDNet是一个Unet结构的双分支网络，它以intensity Map 和HV color map为输入。首先，为了结构的一致性，我们利用3x3的卷积层将intensity map和HV color map变换到相同的维度。然后将其feed到以LCA模块构建的主体结构中。其中LCA包含两路，分别处理图像的亮度增强，图像的噪声抑制和色差矫正。在经过CIDnet的主体结构后，将分别得到处理后的I和HV，我们将其concat在一起并与原始的HVImap相加后作为CIDdual-branch Unet的输出。
Based on the UNet architecture, the enhancement network takes an intensity map and HV color map as input. 
To learn the initial information of intensity map and HV color map, we employ $3\times3$ convolutional layers to obtain the features with same dimension in each branch.
% Initially, for structural consistency, we employ $3\times3$ convolutional layers to transform the intensity map and HV color map to the same dimension. 
Subsequently, the features are fed into the UNet with Lighten Cross-Attention (LCA) modules. The LCA module consists of cross attention blocks, color denoise layer and intensity enhance layer.
The cross attention block learns the corresponding interacted information between the HV-branch and I-branch.
The color denoise layer avoids noise artifacts and color shift, and the intensity enhance layer improves the luminance and removes the saturated regions.
The final outputs of UNet are the refined intensity and HV maps.
To decrease the difficulty of the learning process, we employ a residual mechanism to add the original HVI map.
 % After passing through the backbone structure of CIDNet, processed intensity and HV are obtained separately, concatenated, and added to the original HVI map to serve as the output of the CID dual-branch UNet.
%在得到处理过的HVI空间特征后，我们再执行一个HVI逆变换将图像变换到sRGB空间。这个过程中，首先使用clip操作截断异常值，然后使用公式（？）进行逆变换。同样的，我们使用了可训练参数 density-k，其值与HVI变换保持一致。在经过逆变换后，我们会得到处理后的图像，其不仅亮度得到了增强，而且噪声和色差也被显著的抑制。
Finally, we perform an HVI inverse transformation with the trainable parameter density-$k$ to map the image to the sRGB space.
% After obtaining the processed HVI spatial features, we further perform an HVI inverse transformation to map the image back to the sRGB space. 
% We utilize the trainable parameter density-$k$, whose value remains consistent with the HVI transformation. 
% Thanks to the inverse transformation, the resulting image not only experiences enhanced brightness, but also demonstrates significant suppression of noise and chromatic aberrations.

% The core concept of the CIDNet pipeline is to split an input image into two branches, HV and Intensity, for enhancing and denoising respectively. These branches are later concatenated and transformed into a well-enhanced sRGB image.

% Given a low-light sRGB image,  we first generate an intensity map $\mathbf{I_{I}}=\mathbf{I}_{max}$ using Eq.(\ref{eq:1}), and input the I-map with trainable density-$k$ into HV Transform (HVT) to generate the HV color map using Eq.(\ref{eq:7}) as Fig. \ref{fig:CIDNet-pipeline}.(a). Next, HVIT concatenates two maps to low-light HVI-map $\mathbf{I_{HVI}}$.  I-map and HV-map are embedded by two different $\texttt{Conv}3\times3$ layers and output the I-feature and HV-Feature with shape $H\times W\times C$. 
% Intensity map satisfy $\hat{\mathbf{I}} \in \mathbb{R}^{\mathrm{H\times W}} $, and HV map satisfy $\mathbf{I_{\hat{C}_{H}\hat{C}_{V}}} \in \mathbb{R}^{\mathrm{H\times W\times 2}} $.
\par
% The second step involves the utilization of two specific features as inputs to a CID Dual-branch UNet as Fig.\ref{fig:CIDNet-pipeline}.(b). It outputs the light-up I-feature and denoised HV-feature, both sending to a $\texttt{Conv}3\times3$ layer in each way and concatenating together to a residual HVI map $\mathbf{R_{HVI}} $. There we add the low-light HVI map and residual HVI map as $\hat{\mathbf{I}}=\mathbf{I_{HVI}}+\mathbf{R_{HVI}}$. (
% A detailed exposition of its structure will be provided in Sec. \ref{sec:structure}.) 

\subsection{Structure}
\label{sec:structure}

\subsubsection{Enhancement Network}

%如fig2(b)所展示，我们延续了众多LLIE方法使用的Unet结构，设计了CID Dual-branch UNet，包括encoder，decoder以及多个跳层连接。在我们的网络结构中，encoder部分以LCA为基础，结合了三个downsample和kernel为3的卷积层。同样，decoder也包含了三个LCA模块，同时整合了upsample和卷积层。为了统一处理维度不同的intensity map和HV color map，我们在encoder之前使用了两个卷积层分别用于它们维度的变换，得到维度和大小均相同的intensity feature 和HV feature，并将其输入到encoder中分别处理。
% As depicted in Fig. \ref{fig:CIDNet-pipeline}(b), we extend the utilization of the UNet structure commonly employed in numerous LLIE methods to devise the enhancement network, which comprises an encoder, decoder, and multiple skip connections. 
As depicted in Fig. \ref{fig:CIDNet-pipeline}(b), we use the commonly employed UNet as the baseline network, which has an encoder, decoder, and multiple skip connections. 
The encoder includes three LCA blocks and downsample and $3\times3$ kernel convolutional layers.
Similarly, the decoder consists of three LCA modules, and upsampling layers. 
% To align the dimension of the intensity map and HV color map, two convolutional layers are employed before the encoder for dimension transformation, yielding the intensity feature and HV feature with consistent dimensions and sizes, which are subsequently fed into the encoder for individual processing.

%IEL和CDL也是，主要的写，无关紧要的删掉

% The CID Dual-branch UNet embed six \textcolor[RGB]{255,95,95}{LCA} modules into a network with two U-shape nets as Fig. \ref{fig:CIDNet-pipeline}.(b). It contains an encoder with three DownSample $\texttt{Conv}3\times3$ and a decoder with three UpSample $\texttt{Conv}3\times3$. It inputs the I-branch and HV-branch feature with the shape $H\times W \times C$. Each DownSample reduces $H$ and $W$ by double and doubles the channels. Each UpSample doubles $H$ and $W$ and reduces the channels by double. Finally, it outputs a light-up residual I-branch and HV-branch feature as same as the input tensor shape.\par

\subsubsection{Lighten Cross-Attention}
%encoder和decoder中实现特征处理的主要模块是LCA。为了减少增强图像亮度和抑制图像噪声和色差这两个子任务间的负面影响，我们设计了并行结构的LCA，其中HV-branch和I-branch分别处理HV feature和intensity feature.它们的结构相同，参数一样。此外，处理过程中的HV feature和intensity feature 也存在互补的可能。所以我们提出使用交叉注意力的方式来让HV fearture和intensity feature 来互相引导。由于它们只用于注意力中的query,所以在提供guidance的同时，也隔绝了它们之间的负面交互。
% The primary module responsible for feature processing in the encoder and decoder is the Lighten Cross-Attention (\textcolor[RGB]{255,95,95}{LCA}). 
% To mitigate the negative impact between enhancing image brightness and suppressing image noise and chromatic aberrations, we designed a parallel-structured LCA as Fig. \ref{fig:CIDNet-structure}, where the HV-branch and I-branch handle HV features and intensity features, respectively. They share the same structure and parameters.
To enhance the interaction between the structures of images contained in the brightness and color branches, we propose the Lighten Cross-Attention (LCA) module to learn the complementary information of HV-branch and intensity-branch.
As shown in Fig. \ref{fig:CIDNet-structure}, the HV-branch and I-branch in LCA handle HV features and intensity features, respectively. 
To learn the complementary potential between HV features and intensity features during the processing, we propose cross attention block (\textbf{CAB}) to facilitate mutual guidance between HV features and intensity features.
To force the CAB to learn the information from the opposite branch (\emph{i.e.}, HV branch only use the information of I-brach to refine itself), we utilize the one branch as the query and leverage another branch as key and value in the CAB.

\begin{figure*}
    \centering
    \includegraphics[width=1\linewidth]{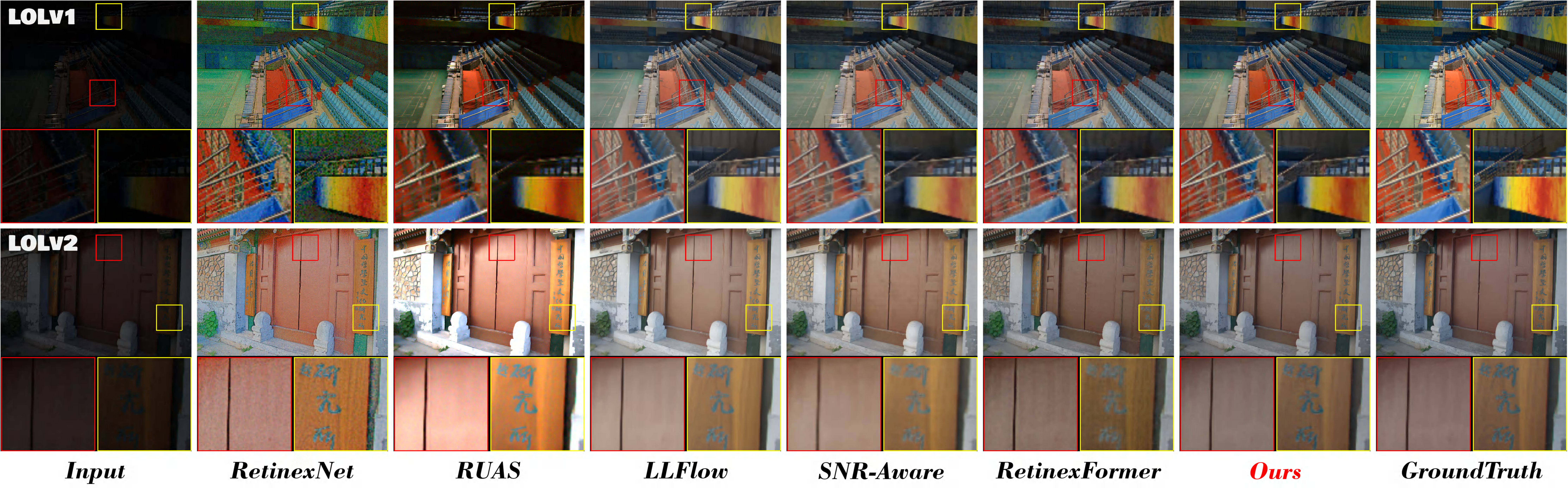}
    \caption{Visual comparisons of the enhanced results by different methods on LOLv1 and LOLv2. (\textbf{Zoom in for the best view.}) }
    \label{fig:LOL}
\end{figure*}
% The color denoise layer avoids noise artifacts and color shift, and the intensity enhance layer improves the luminance and removes the saturated regions.
%  Since they are only used as queries in the attention mechanism, the cross-attention not only provides guidance but also isolates negative interactions between them.

%详细来说， **********（可以把公式（10）这一段改改放到这里，符号不要太多，rearrange这变到什么大小了可以不说，主要的留）
As shown in \ref{fig:CIDNet-structure}, 
% Specifically, the LCA module consists of a Cross Attention Block (\textbf{CAB}), an Intensity Enhance Layer (\textbf{IEL}) for the I-way, and a Color Denoise Layer (\textbf{CDL}) for the HV-way. 
The CAB exhibits a symmetrical structure between the I-way and HV-way. We use the I-branch as an example to describe the details.
% , while the design of IEL and CDL is identical but aligns with different theoretical principles. For ease of explanation, we will focus on the process within the I-way as follows.
$\mathbf{Y_{I}}\in \mathbb{R}^{\hat{H}\times\hat{W}\times\hat{C}}$ denotes the inputs of I-branch, our CAB first derives query ($\mathbf{Q}$)  by  $\mathbf{Q}=W^{(Q)}\mathbf{Y_{I}}$. Meanwhile, the CAB splits key ($\mathbf{K}$) and value ($\mathbf{V}$) by $\mathbf{K}=W^{(K)}\mathbf{Y_{I}}$ and $\mathbf{V}=W^{(V)}\mathbf{Y_{I}}$. $W^{(Q)}$, $W^{(K)}$ and $W^{(V)}$ represents the feature embedding convolution layers. We formulate as 
\begin{equation}
\begin{split}
\mathbf{\hat{Y}_I}=W(\mathbf{V} \otimes \mathrm{Softmax}\left( \mathbf{Q}\otimes \mathbf{K}/\alpha_H\right) + \mathbf{Y_I})
\end{split}
\label{eq:11}
\end{equation}
where $\alpha_H$ is the multi-head factor \cite{dosovitskiy2021image} and $W(\cdot)$ denotes the feature embedding convolutions.

Next, following Retinex theory, intensity enhance layer (IEL) decomposes the tensor $\mathbf{\hat{Y}_{I}}$ as $\mathbf{Y}_I=W^{(I)}\mathbf{\hat{Y}_{I}}$ and $\mathbf{Y}_R=W^{(R)}\mathbf{\hat{Y}_{I}}$. The IEL is defined as 
\begin{equation}
\begin{split}
\mathbf{\tilde{Y}_{I}} = W_{s}(&(\tanh{(W_{s}\mathbf{Y}_I)} + \mathbf{Y}_I) \\ \odot&(\tanh{(W_{s}\mathbf{Y}_R)} + \mathbf{Y}_R))
\end{split}
\label{eq:12}
\end{equation}
where $\odot$ represents the element-wise multiplication and $W_{s}$ denotes the depth-wise convolution layers. Finally, the output of IEL adds the residuals to simplify the training process. 
% The key disparity in structure between the HV-way and I-way lies in the absence of residual processing in the CDL, a deliberate choice made to adhere to the principles of color optical decomposition theory \cite{BORN1980133}.

% The IEL has the same structure as CDL, but to reduce the training difficulty, IEL outputs the residuals as $\mathbf{\tilde{Y}_{I}}=\mathrm{IEL}(\mathbf{\hat{Y}'_{I}})+\mathbf{\hat{Y}_{I}}$.

% For the purpose of reduce the computational overhead of the Dual-branch UNet, we specially designed the Lighten Cross-Attention (\textcolor[RGB]{255,95,95}{LCA}) module which only has linear complexity and well-controlled color features based on Transformer \cite{petit2021unet} as Fig. \ref{fig:CIDNet-structure}.\par

% \textbf{CAB.} \par

\subsection{Loss Function}
% Given an output HVI map $\mathbf{\hat{I}_{HVI}}$ and a sRGB image $\mathbf{\hat{I}}$. Let $\mathbf{I}$ represent sRGB GroundTruth, which can be transformed to HVI map $\mathbf{I_{HVI}}$. We employ L1 loss $L_1$, edge loss $L_e$ \cite{Edgeloss} and perceptual loss $L_p$ \cite{johnson2016perceptual} at sRGB and HVI space as
% \begin{equation}
% \begin{split}
% &\ l(\hat{x},x)=\lambda_{1} \cdot L_1(\hat{x},x) + \lambda_{e} \cdot L_e+ \lambda_{p} \cdot L_p(\hat{x},x)\\
% &\ L= \lambda_{c}\cdot l(\mathbf{\hat{I}_{HVI}},\mathbf{I_{HVI}}) + l(\mathbf{\hat{I}},\mathbf{I})\\
% \end{split}
% \label{eq:14}
% \end{equation}
% where $\lambda_{c},\lambda_{1},\lambda_{e},\lambda_{p}$ are all the weight coefficients used to trade-off the loss function $L$ from our experiments.

%我们的损失分为sRGB空间的损失和HVI空间的损失，（公式12分开，不要上下两行）。在每个空间中，我们使用了增强任务中常用的L1,edge loss何perceptual loss,其可以写为：公式（12）
%虽然我们提出了HVI空间，并在此空间中进行LLIE的处理，但是在我们的HVI变换和逆变换中有着可训练的参数，所以我们也使用了sRGB空间的损失函数，不仅能优化k等参数，也能起到正则化的作用。所以，我们总的损失函数为 公式（13）
To integrate the advantages of HVI  space and the sRGB space, the loss function consists both color spaces.
% The loss function consists of losses in the sRGB space and the HVI space. 
In HVI color space, we utilize L1 loss $L_1$, edge loss $L_e$ \cite{Edgeloss}, and perceptual loss $L_p$ \cite{johnson2016perceptual} for the low-light enhancement task. It can be expressed as 
\begin{equation}
\begin{split}
l(\hat{X}_{HVI},X_{HVI})&=\lambda_{1} \cdot L_1(\hat{X}_{HVI},X_{HVI}) \\
&+ \lambda_{e} \cdot L_e(\hat{X}_{HVI},X_{HVI})\\
&+ \lambda_{p} \cdot L_p(\hat{X}_{HVI},X_{HVI})
\end{split}
,
\label{eq:14}
\end{equation}
where $\lambda_{1},\lambda_{e},\lambda_{p}$ are all the weight to trade-off the loss function $l(\cdot)$.
In sRGB color space, we employ the same loss function as $l(\hat{I}, I)$.
% Despite introducing the HVI space, the trainable parameters in our HVI transformation and inverse transformation lead us to employ loss functions in the sRGB space as well. This not only optimizes parameters like $k$ but also serves as a regularization term. 
Therefore, our overall loss function $L$ is represented by 
\begin{equation}
L= \lambda_{c}\cdot l(\mathbf{\hat{I}_{HVI}},\mathbf{I_{HVI}}) + l(\mathbf{\hat{I}},\mathbf{I})
\end{equation}
where $\lambda_{c}$ is the weight to balance the loss in different color spaces.

\section{Experiments}

\subsection{Datasets and Settings}
We employ seven commonly-used LLIE benchmark datasets for evaluation, including LOLv1 \cite{RetinexNet}, LOLv2 \cite{LOLv2}, DICM \cite{DICM}, LIME \cite{LIME}, MEF \cite{MEF}, NPE \cite{NPE}, and VV \cite{VV}. We also conduct further experiments on two extreme datasets, SICE \cite{SICE} (containing mix and grad test sets \cite{SICE-Mix}) and SID (Sony-Total-Dark) \cite{SID}. Since blurring is often prone to occur in low-luminosity images, to demonstrate the robustness of our CIDNet to multitasking, we conducted experiments on LOL-Blur \cite{LEDNet} as well.\par

\textbf{LOL.} The LOL dataset has v1 \cite{RetinexNet} and v2 \cite{LOLv2} versions. LOL-v2 is divided into real and synthetic subsets. The training and testing sets are split in proportion to 485:15, 689:100, and 900:100 on LOL-v1, LOL-v2-Real, and LOL-v2-Synthetic. For LOLv1 and LOLv2-Real, we crop the training images into $400\times400$ patches and train CIDNet for 1500 epochs with a batch size of 8. For LOLv2-Synthetic, we set the batch size to 1 and trained 500 epochs without cropping.

\textbf{SICE.} The original SICE dataset \cite{SICE} contains a total of 589 sets of low-light and overexposed images, and the training set, validation set, and test set are divided into three groups according to 7:1:2. We train on the SICE training set with the batch size of 10 and test on the datasets SICE-Mix and SICE-Grad \cite{SICE-Mix}. We crop the original SICE image by $160\times160$ and train CIDNet over 1000 epochs.\par

\begin{figure}
    \centering
    \includegraphics[width=1.0\linewidth]{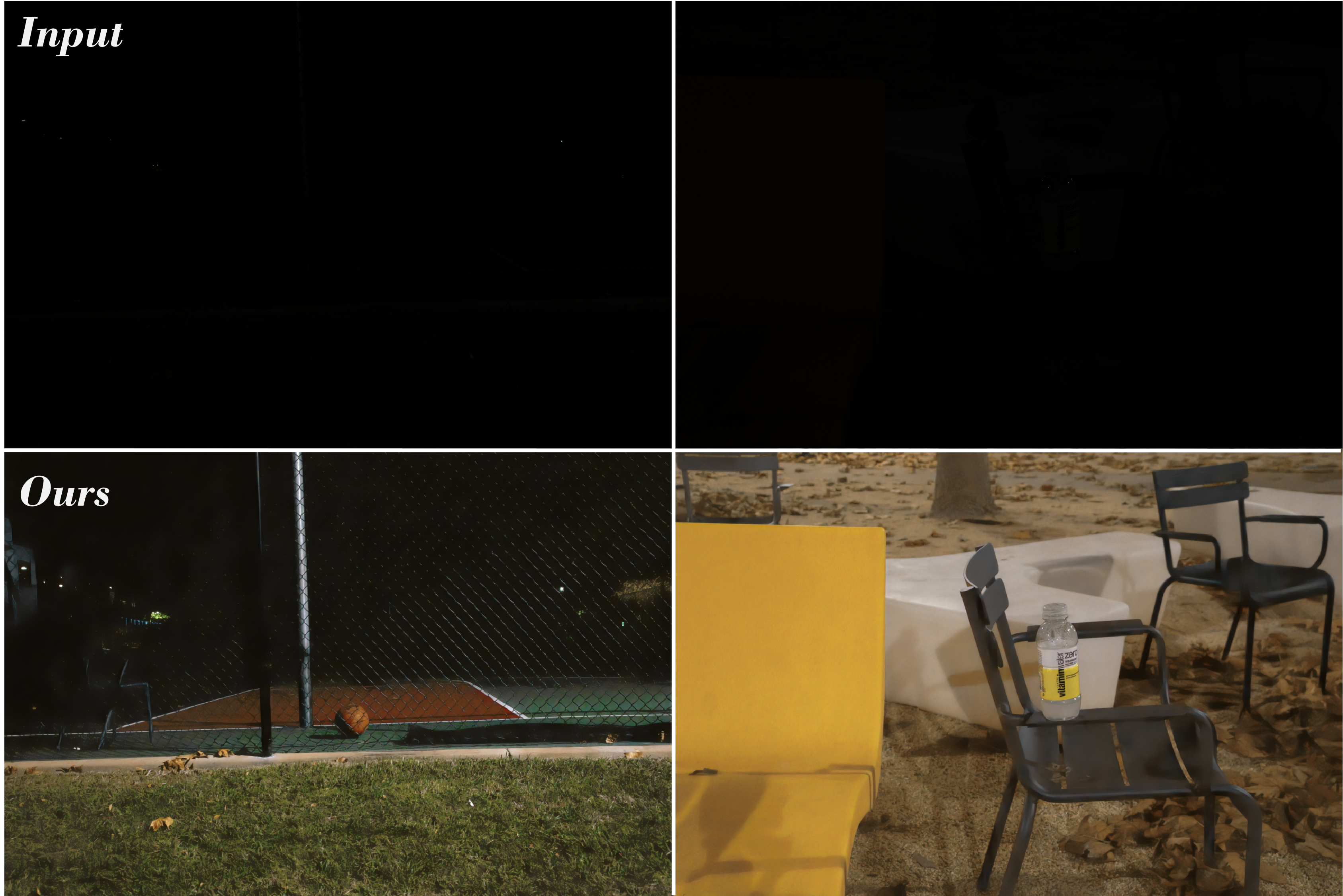}
    \caption{Input Sony-Total-Dark extreme low-light image with the image enhanced by our CIDNet.}
    \label{fig:SID-show}
\end{figure}
\begin{table*}
    \centering
    \renewcommand{\arraystretch}{1.15}
    \caption{Quantitative comparisons PSNR/SSIM$\uparrow$ on LOL (v1 and v2) datasets. Normal and GT Mean represent with and without gamma-corrected by GroundTruth. The highest result is in\textcolor{red}{~red} color, the second highest result is in\textcolor{cyan}{~cyan} color, and the third is in \textcolor{green}{green}. wP and oP represent to train on CIDNet with and without perceptual loss \protect\cite{VGG19}.}
    \resizebox{\textwidth}{!}{
    \begin{tabular}{c|cc|cccc|cccc|cccc}
        \Xhline{1.5pt}
        \multirow{3}{*}{\textbf{Methods}}&\multicolumn{2}{c|}{\multirow{2}{*}{\textbf{Complexity}}}&\multicolumn{4}{c|}{\textbf{LOLv1}} & \multicolumn{4}{c|}{\textbf{LOLv2-Real}} & \multicolumn{4}{c}{\textbf{LOLv2-Syn}}\\
        
        ~&~&~&\multicolumn{2}{c}{Normal} & \multicolumn{2}{c|}{GT~Mean} & \multicolumn{2}{c}{Normal} & \multicolumn{2}{c|}{GT~Mean}& \multicolumn{2}{c}{Normal} & \multicolumn{2}{c}{GT~Mean}\\
        \cline{2-15}
        ~&Params/M&FLOPs/G&PSNR&SSIM&PSNR&SSIM&PSNR&SSIM&PSNR&SSIM&PSNR&SSIM&PSNR&SSIM\\
        \hline
        RetinexNet \cite{RetinexNet}&  0.84& 584.47 &16.774 &	0.419 &	18.915 &	0.427 &	16.097 &	0.401 	&18.323 &	0.447 &	17.137 &	0.762 &	19.099& 	0.774 
\\
        KinD \cite{KinD}&  8.02& 34.99 &17.650 &	0.775 &	20.860& 	0.802 &	14.740& 	0.641 &	17.544 	&0.669 &	13.290 	&0.578 &	16.259 &	0.591 
\\
         ZeroDCE \cite{Zero-DCE}& 0.075&4.83& 14.861 &	0.559 &	21.880 &	0.640& 16.059 &	0.580 &	19.771 &	0.671 &	17.712& 	0.815 &	21.463 &	0.848 
\\
         3DLUT \cite{3DLUT}&  0.59& 7.67 & 14.350 & 	0.445 & 	21.350&  	0.585 & 	17.590&  	0.721 & 	20.190 & 	0.745 & 	18.040 & 	0.800 & 	22.173 & 	0.854 
\\
        DRBN \cite{DRBN}& 5.47 & 48.61 & 16.290 & 0.617 & 19.550 & 0.746 & 20.290 & 0.831 & - & -&23.220 &0.927 & -&-
         
\\
%         MIRNet \cite{MIRNet}& 31.76 & 785.1 &24.100 &	0.845 &	26.519 &	0.856 &	20.020 &	0.820& 	27.173 &	0.865 &	21.940 &	0.876 &	25.955& 	0.898 
% \\
         RUAS \cite{RUAS}&  0.003& 0.83 &16.405 &	0.500 &	18.654& 	0.518 &	15.326 &	0.488 &	19.061 &	0.510 	&13.765 	&0.638 	&16.584 &	0.719 
\\
        LLFlow \cite{LLFlow}& 17.42 & 358.4 &21.149 &	\textcolor{green}{0.854} 	&24.998 &	\textcolor{green}{0.871} &	17.433 &	0.831 &	25.421 	&\textcolor{green}{0.877}&	24.807 &	0.919 &	27.961 &	0.930 
\\
         EnlightenGAN \cite{EnGAN}& 114.35 & 61.01 & 17.480 & 0.651 & 20.003 & 0.691 & 18.230 & 0.617 & - &- & 16.570& 0.734& -&-
\\

         Restormer \cite{Restormer}& 26.13 & 144.25 &22.365 	&0.816 &	26.682 	&0.853 	&18.693 &	0.834 &	26.116& 	0.853 &	21.413 &	0.830 &	25.428 &	0.859 
\\

         LEDNet \cite{LEDNet}& 7.07 & 35.92 &20.627 &	0.823 &	25.470 &	0.846 &	19.938 &	0.827 &	\color{cyan}{27.814} &	0.870 &	23.709 &	0.914 &	27.367 &	0.928 
\\
        
         SNR-Aware \cite{SNR-Aware}& 4.01 & 26.35 &\color{cyan}{24.610} 	&0.842 	&26.716 	&0.851 	&21.480 &	\textcolor{green}{0.849} &	27.209 &	0.871 &	24.140 &	0.928 &	27.787 &	\textcolor{green}{0.941} 
\\
        PairLIE \cite{PairLIE}& 0.33 & 20.81 &19.510 &	0.736 &	23.526 &	0.755 &	19.885 &	0.778 &	24.025 &	0.803 &	-	&-&	-&	-
\\
        LLFormer \cite{LLFormer}&24.55&22.52&23.649&0.816&25.758&0.823&20.056&0.792&26.197 &0.819&24.038&0.909&28.006& 0.927
\\
         RetinexFormer \cite{RetinexFormer}& 1.53 & 15.85 & \color{red}{25.153}  &	0.846  &	\textcolor{green}{27.140} & 	0.850  &	\textcolor{green}{22.794}  &	0.840  &	27.694  &	0.856  &	\color{cyan}{25.670}  &	\textcolor{green}{0.930}  &	\textcolor{green}{28.992} & 	0.939 
\\
         \Xhline{1.5pt}
        \textbf{CIDNet-wP}& 1.88 & 7.57 & \textcolor{green}{23.809} &	\color{cyan}{0.857} &	\color{cyan}{27.715} &	\color{cyan}{0.876} &	\color{red}{24.111}& 	\color{red}{0.868} &	\color{red}{28.134}& 	\color{red}{0.892} &	\textcolor{green}{25.129} &	\color{cyan}{0.939} &	\color{cyan}{29.367}& 	\color{red}{0.950 }
\\
         \textbf{CIDNet-oP}& 1.88 & 7.57 &23.500 &	\color{red}{0.870} &	\color{red}{28.141 }&	\color{red}{0.889} &	\color{cyan}{23.427}& 	\color{cyan}{0.862} &	\textcolor{green}{27.762}& 	\color{cyan}{0.881}& 	\color{red}{25.705}& 	\color{red}{0.942}& 	\color{red}{29.566}& 	\color{red}{0.950 }
\\
         \Xhline{1.5pt}
    \end{tabular}
    }
    \label{tab:table-LOL}
\end{table*}
 \begin{figure*}
    \centering
    \includegraphics[width=1\linewidth]{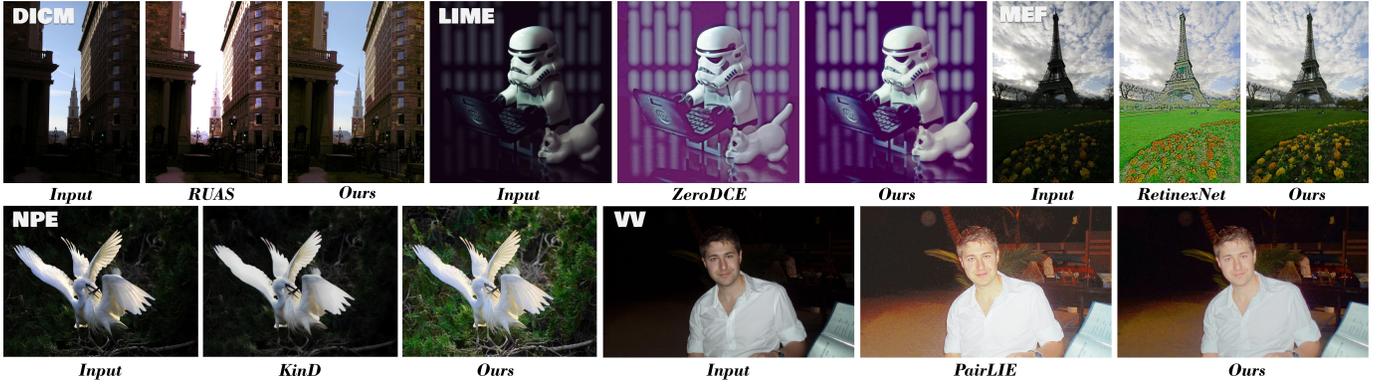}
    \caption{Five unpaired datasets are compared visually, and we randomly select one image in each dataset to compare with the other methods. Our CIDNet enhances dark details and illumination to a suitable interval, which is better than the other methods.}
    \label{fig:unpaired}
\end{figure*}

\textbf{Sony-Total-Dark.} This dataset is a customized version of the subset of SID dataset captured by the Sony $\alpha$7S II camera, which is adopted for evaluation. There are 2697 short-long-exposure RAW image pairs. To make this dataset more challenging, we convert the RAW format images to sRGB images with \textit{no gamma correction} as Fig. \ref{fig:SID-show}, which resulted in images becoming extremely dark. We crop the training images into $256\times256$ patches and train CIDNet for 1500 epochs with a batch size of 4.\par

\textbf{LOL-Blur.} The dataset, LOL-Blur \cite{LEDNet}, contains 12,000 low-blur/normal-sharp pairs with diverse darkness and blurs in 200 different scenarios.  The training and testing sets are split in proportion to 17:3. In our experiment, we use low-blur and high-sharp-scaled sets for training and testing. We crop the training images into $256\times256$ patches and train CIDNet for 300 epochs with a batch size of 8. Unlike other experiments, here we use $l_{mse}$ Loss instead of $l_1$ Loss in Eq. \ref{eq:14}.\par

\textbf{Experiment Settings.} We implement our CIDNet by PyTorch. The model is trained with the Adam \cite{Adam} optimizer (\textit{$\beta_{1}$} = 0.9 and \textit{$\beta_{2}$} = 0.999) for at least 300 epochs by using a single NIVIDA 2080Ti or 3090 GPU. The learning rate is initially set to $1 \times 10^{-4}$ and then steadily decreased to $1 \times 10^{-7}$ by the cosine annealing scheme \cite{sgdr} during the training process. In testing stage, we pad the input images to be a multiplier of $8\times8$ using reflect padding on both sides. After inference, we crop the padded image back to its original size. Since the outliers are existed in the outputs of UNets, we simply use clip operation to HVI space, which follows $\mathbb{D}=\{p=(h,v,i)|~h^2 + v^2 \le \sin^\frac{2}{k}(\frac{\pi i}{2}),~0 \le i \le 1\}$, where $p=(h,v,i)$ is the three-dimensional coordinates in HVI color space, and $k$ denotes the density-$k$ in Eq. \ref{eq:2}.

\begin{table}
    \centering
    \renewcommand{\arraystretch}{1.2}
    \caption{Quantitative comparisons LPIPS/FLOPs$\downarrow$ on LOL (v1 and v2) datasets. GT Mean represents the gamma-corrected image by GroundTruth. The best result is in \textcolor{red}{red} color.}
    \resizebox{\linewidth}{!}{
    \begin{tabular}{c|cc|cc|cc|c}
     \Xhline{1.5pt}
         \multirow{2}{*}{\textbf{Methods}}&
         \multicolumn{2}{c|}{\textbf{LOLv1}}& 
         \multicolumn{2}{c|}{\textbf{LOLv2-Real}} & 
         \multicolumn{2}{c|}{\textbf{LOLv2-Syn}}&\textbf{Complexity}\\
         ~& Normal & GT~Mean & Normal & GT~Mean &  Normal & GT~Mean & FLOPs/G$\downarrow$ \\
         \hline
         EnlightenGAN & 0.322 &	0.317 &	0.309 &	0.301 &	0.220 &	0.213 & 61.01
\\
         RetinexNet &0.474 &	0.470 	&0.543 &	0.519 &	0.255& 	0.247 &	587.47
\\
        LLFormer &0.175& 	0.167 &	0.211 &	0.209& 	0.066 &	0.061 &	22.52
\\
        LLFlow &0.119 &	0.117 &	0.176 &	0.158 &	0.067 &	0.063 &358.4
\\
         LEDNet &0.118&	0.113&	0.120 &	0.114&	0.061&	0.056 & 35.92
\\
        RetinexFormer &0.131 &	0.129 &	0.171 &	0.166 &	0.059 &	0.056 &15.85
\\
          \Xhline{1.5pt}
         \textbf{CIDNet}&\color{red}0.086 &	\color{red}{0.079} &	\color{red}{0.108} &	\color{red}{0.101} &	\color{red}{0.045} 	&\color{red}{0.040} &	\color{red}{7.57}
\\
          \Xhline{1.5pt}
    \end{tabular}
}
    
    \label{tab:table-LPIPS}
\end{table}
\vspace{-0.2 pt}

\textbf{Evaluation Metrics. }For the \textit{paired} dataset, we adopt the Peak Signal-to-Noise Ratio (PSNR) and Structural Similarity (SSIM) \cite{SSIM} as the distortion metrics. 
% Unfortunately, PSNR is well-known to only partially correspond to human perception and can lead to algorithms with visibly lower quality in the reconstructed images. 
To comprehensively evaluate the perceptual quality of restored images, we report Learned Perceptual Image Patch Similarity (LPIPS) \cite{LPIPS} by using AlexNet \cite{Alex} for references as a perceptual metric. For the \textit{unpaired} datasets, we evaluated single recovered images using BRISQUE \cite{BRISQUE} and NIQE \cite{NIQE} perceptually.

 \begin{table*}[htbp]
      \centering
        \renewcommand{\arraystretch}{1.2}
        \caption{Quantitative comparison on five unpaired datasets with \textbf{BRISQUE}$\downarrow$ and \textbf{NIQE}$\downarrow$. The best result is in \textcolor{red}{red} color.}
        \resizebox{\linewidth}{!}{
        \begin{tabular}{c|cc|cc|cc|cc|cc}
        \Xhline{1.5pt}
        \multirow{2}{*}{\textbf{Methods}}&
         \multicolumn{2}{c|}{\textbf{DICM}}& 
         \multicolumn{2}{c|}{\textbf{LIME}} & 
         \multicolumn{2}{c|}{\textbf{MEF}} & 
         \multicolumn{2}{c|}{\textbf{NPE}} & 
         \multicolumn{2}{c}{\textbf{VV}}\\

        ~&
            BRISQUE$\downarrow$&	NIQE$\downarrow$&
            BRISQUE$\downarrow$&	NIQE$\downarrow$&
            BRISQUE$\downarrow$&	NIQE$\downarrow$&
            BRISQUE$\downarrow$&	NIQE$\downarrow$&
            BRISQUE$\downarrow$&	NIQE$\downarrow$
            \\
            
            \hline
            KinD \cite{KinD}&
            48.72& 	5.15& 	
            39.91& 	5.03& 	
            49.94& 	5.47& 	
            36.85& 	4.98& 	
            50.56& 	4.30

\\
            ZeroDCE \cite{Zero-DCE}&
            27.56& 	4.58 &
            20.44& 	5.82 &
            17.32& 	4.93 &
            20.72& 	4.53 &
            34.66& 	4.81  
\\
                 
            RUAS \cite{RUAS}&
            38.75& 	5.21& 	
            27.59& 	4.26& 	
            23.68& 	3.83& 	
            47.85& 	5.53& 	
            38.37& 	4.29 

\\
            LLFlow \cite{LLFlow}&
            26.36& 	4.06& 
            27.06& 	4.59& 
            30.27& 	4.70& 
            28.86& 	4.67& 
            31.67& 	4.04 
\\
            SNR-Aware \cite{SNR-Aware}&
            37.35& 	4.71& 
            39.22& 	5.74& 	
            31.28& 	4.18& 	
            26.65& 	4.32& 	
            78.72& 	9.87

\\
            PairLIE \cite{PairLIE}&
            33.31& 	4.03& 	
            25.23& 	4.58& 	
            27.53& 	4.06& 	
            28.27& 	4.18& 	
            39.13& 	3.57

\\
             \Xhline{1.5pt}
             \textbf{CIDNet}&
            \color{red}{21.47}&	\color{red}{3.79}&	
            \color{red}{16.25}& \color{red}{4.13}&
            \color{red}{13.77}&	\color{red}{3.56}&
            \color{red}{18.92}&	\color{red}{3.74}&	
            \color{red}{30.63}& \color{red}{3.21}\\
            \Xhline{1.5pt}
        \end{tabular}
        }
        
        \label{tab:unpaired}
    \end{table*}%

 \begin{table}[b]
      \centering
        \renewcommand{\arraystretch}{1.1}
        \caption{Quantitative comparison on three hard datasets (SICE-Mix/Grad and Sony-Total-Dark) with \textbf{PSNR}$\uparrow$, \textbf{SSIM}$\uparrow$ and \textbf{LPIPS}$\downarrow$.}
        \resizebox{\linewidth}{!}{
        \begin{tabular}{c|ccc|ccc|ccc}
        \Xhline{1.5pt}
        \multirow{2}{*}{\textbf{Methods}}&
         \multicolumn{3}{c|}{\textbf{SICE-Mix}}& 
         \multicolumn{3}{c|}{\textbf{SICE-Grad}} & 
         \multicolumn{3}{c}{\textbf{Sony-Total-Dark}}\\

        ~&
            PSNR$\uparrow$&	SSIM$\uparrow$&	LPIPS$\downarrow$& 
            PSNR$\uparrow$&	SSIM$\uparrow$&	LPIPS$\downarrow$&
            PSNR$\uparrow$&	SSIM$\uparrow$&	LPIPS$\downarrow$\\
            
            \hline

            RetinexNet&
            12.397& 0.606& 	0.407& 
            12.450& 0.619& 	0.364& 
            15.695& 0.395& 	0.743 
\\
            ZeroDCE&
            12.428& 0.633& 0.382& 
            12.475& 0.644& 0.334& 
            14.087&	0.090& 	0.813
\\
            URetinexNet&
            10.903& 0.600& 	0.402& 
            10.894& 0.610& 	0.356& 
            15.519& 0.323& 	0.599 
\\
            RUAS&
            8.684& 	0.493& 	0.525& 
            8.628& 	0.494& 	0.499& 
            12.622& 0.081& 	0.920 
\\
            LLFlow&
            12.737&	0.617& 	0.388& 
            12.737& 0.617& 	0.388& 
            16.226&	0.367& 	0.619 
\\
            LEDNet&
            12.668&	0.579&	0.412& 
            12.551& 0.576& 	0.383& 
            20.830& 0.648& 	0.471 
\\

             \Xhline{1.5pt}
             \textbf{CIDNet}&
            \color{red}{13.425}&	\color{red}{0.636}&	\color{red}{0.362}& 
            \color{red}{13.446}&	\color{red}{0.648}&	\color{red}{0.318}&
            \color{red}{22.904}&	\color{red}{0.676}&	\color{red}{0.411}\\
            \Xhline{1.5pt}
        \end{tabular}
        }
        
        \label{tab:SID}
    \end{table}

\subsection{Evaluation on Image Enhancement}

\textbf{LOL Datasets Results.} 
We quantitatively compare our CIDNet with many SOTA methods as shown in Table \ref{tab:table-LOL} and \ref{tab:table-LPIPS}. It can be found that our method is optimal on almost all metrics for both LOLv1 and LOLv2 datasets. Comparing the two tables comprehensively, it can be seen that CIDNet achieves six new SOTA SSIM and LPIPS results on three subsets of LOL (v1 and v2) dataset with \textcolor{red}{only 7.57 GFLOPs}. It outperforms the current SOTA method Retinexformer in terms of both PSNR and SSIM under GT Mean while LPIPS improves by \textcolor{red}{38\%}, \textcolor{red}{39\%}, and \textcolor{red}{29\%}, respectively, and FLOPs decreased by \textcolor{red}{52\%}. Compared to 3DLUT with about the same size of FLOPs, our method significantly improves the PSNR by \textcolor{red}{6.791}, \textcolor{red}{7.944}, and \textcolor{red}{7.393 dB}. It may be observed in Figure \ref{fig:LOL} that our model restores colors extremely well, which may be attributed to the HVI color space.

% Even though our normal PSNR on the LOLv1 dataset is lower compared to some existing methods, it should be noted that a higher PSNR method typically tends to introduce more noise and result in inferior visualization. Hence, our experiment efforts are directed towards enhancing SSIM and LPIPS metrics, which are considered more effective and reliable measures for evaluating image quality.\par

% \begin{figure*}[htbp]
%     \centering
%     \includegraphics[width=1\linewidth]{img/patch2.png}
%     \caption{ Visual comparisons of the enhanced results by different methods on SID-Total-Dark dataset. (\textbf{Zoom in for the best view.})}
%     \label{fig:hard}
% \end{figure*}
\textbf{SICE and Sony-total-Dark.} To verify that our model also performs well on large-scale datasets, we conducted experiments on two extremely difficult-to-recover datasets, SICE (including Mix and Grad) and SID-Total-Dark. The three metrics of our CIDNet \textbf{are optimal on all three test sets} as Table \ref{tab:SID}. especially on the Sony-Total-Dark dataset, which outperforms LEDNet by \textcolor{red}{9.96\%} for the PSNR, \textcolor{red}{4.32\%} for the SSIM, \textcolor{red}{12.74\%} for the LPIPS metrics.

\begin{table}[b]
\centering
\captionof{table}{Ablation of CIDNet module designs. The first row is a simple UNet baseline \cite{petit2021unet} without attention module. The row (2), (3), (4), (6), and (7) (where the rows with LCA but without CrossAttn) use Self-Attention \cite{Restormer} to replace Cross-Attention.}
\label{tab:ablation-a}
\vspace{-1.8mm}
\renewcommand{\arraystretch}{1.1}
\resizebox{\linewidth}{!}{
    \begin{tabular}{ccccc|cc}
    \Xhline{1.5pt}
         \cellcolor{gray!10}&
         \cellcolor{gray!10}Color Space&
         \cellcolor{gray!10}LCA&		\cellcolor{gray!10}Dual-branch&	\cellcolor{gray!10}CrossAttn&	\cellcolor{gray!10}PSNR$\uparrow$&	\cellcolor{gray!10}SSIM$\uparrow$
\\
    \Xhline{1.5pt}
         1&sRGB&	&	&		&16.518& 	0.721
\\
         2&sRGB&$\surd$&	&		&18.606& 	0.822
\\
         3&HSV&$\surd$	&	& &13.237 & 0.365		
\\
         4&HSV&$\surd$	&$\surd$	& &10.236  &0.254 	
\\
         5&HSV&$\surd$	&$\surd$	&$\surd$& 13.668& 0.407	
\\
         6&HVI&$\surd$	&  	&  &22.000&	0.853
\\

         7&HVI&$\surd$	&$\surd$	&   &23.159& 	0.854
\\
         8&HVI&$\surd$	&$\surd$	&$\surd$&\textcolor{red}{24.111}& 	\textcolor{red}{0.868}\\

    \Xhline{1.5pt}
    \end{tabular}}
\end{table}

\textbf{Unpaired Datasets Experiments.} In the case of unpaired datasets DICM, LIME, MEF, NPE, and VV, where GroundTruth is unavailable, we evaluate the effectiveness of models trained on LOLv1 or LOLv2-Syn using various methods, and measure their performance using BRISQUE and NIQE metrics. We report our comparisons against SOTA methods as Table \ref{tab:unpaired}, where our method outperformed all previous SOTA methods. Notably, our method exhibits a substantial improvement in the NIQE metric compared to other approaches.\par
 For each of these five unpaired datasets, we randomly selected an image in each dataset and compared it visually. As Fig. \ref{fig:unpaired}, CIDNet improves the brightness and increases the perceived level of the image while ensuring reasonable color accuracy compared with RUAS, ZeroDCE, RetinexNet, KinD and PairLIE.

% \begin{figure*}
% \centering 
% \subfloat[\centering{Input\\ PSNR$\uparrow$/SSIM$\uparrow$}]{\includegraphics[width=0.15\textwidth]{ablation/input.png}}
% \quad
% \subfloat[\centering{w/o CAB\\ $14.15/0.843$}]{\includegraphics[width=0.15\textwidth]{ablation/0.png}}
% \quad
% \subfloat[\centering{w/o IEL\\ $14.55/0.710$}]{\includegraphics[width=0.15\textwidth]{ablation/1.png}}
% \quad
% \subfloat[\centering{w/o CDL\\ $13.71/0.657$}]{\includegraphics[width=0.15\textwidth]{ablation/2.png}}
% \quad
% \subfloat[\centering{Full LCA\\ $20.80/0.848$}]{\includegraphics[width=0.15\textwidth]{ablation/3.png}}
% \quad
% \subfloat[\centering{GroundTruth\\ $\infty/1.0$}]{\includegraphics[width=0.15\textwidth]{ablation/gt.png}}
% \caption{The visual quality comparison results on LOLv2-Real dataset with various LCA blocks (by removing submodules in the LCA). (e) Full LCA denotes the original design of the LCA block.} 

% \label{fi3}
% \end{figure*}

\begin{figure*}
\centering 
\subfloat[\centering Input $\text{PSNR}\uparrow/\text{SSIM}\uparrow$]{\includegraphics[width=0.15\textwidth]{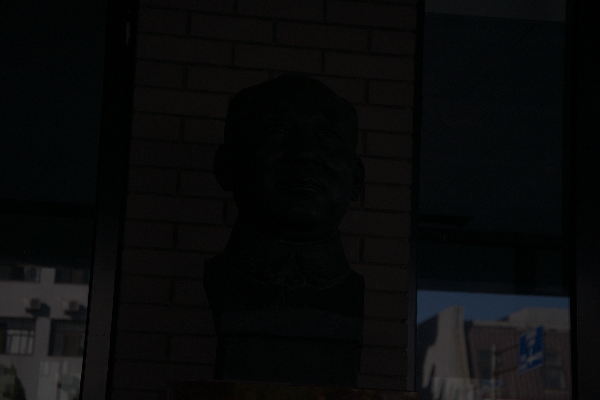}}
\quad
\subfloat[\centering w/o CAB $14.15/0.843$]{\includegraphics[width=0.15\textwidth]{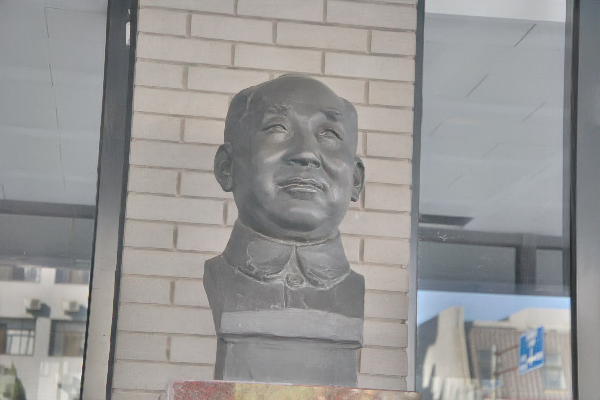}}
\quad
\subfloat[\centering w/o IEL $14.55/0.710$]{\includegraphics[width=0.15\textwidth]{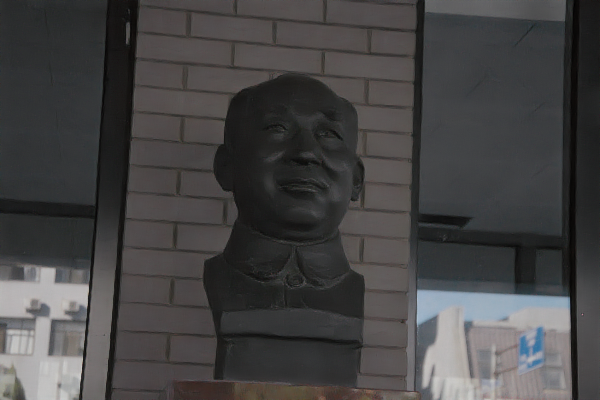}}
\quad
\subfloat[\centering w/o CDL $13.71/0.657$]{\includegraphics[width=0.15\textwidth]{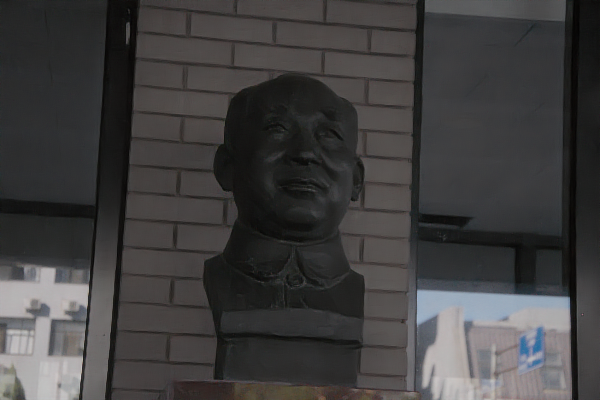}}
\quad
\subfloat[\centering Full LCA $20.80/0.848$]{\includegraphics[width=0.15\textwidth]{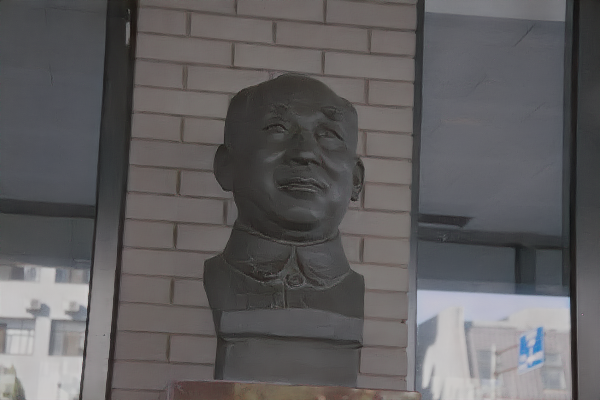}}
\quad
\subfloat[\centering GroundTruth $\infty/1.0$]{\includegraphics[width=0.15\textwidth]{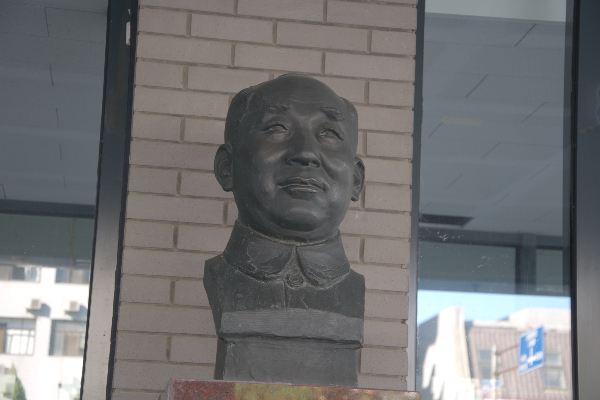}}
\caption{The visual quality comparison results on LOLv2-Real dataset with various LCA blocks (by removing submodules in the LCA). (e) Full LCA denotes the original design of the LCA block.} 
\label{fi3}
\end{figure*}

\begin{figure}
    \centering
    \includegraphics[width=1\linewidth]{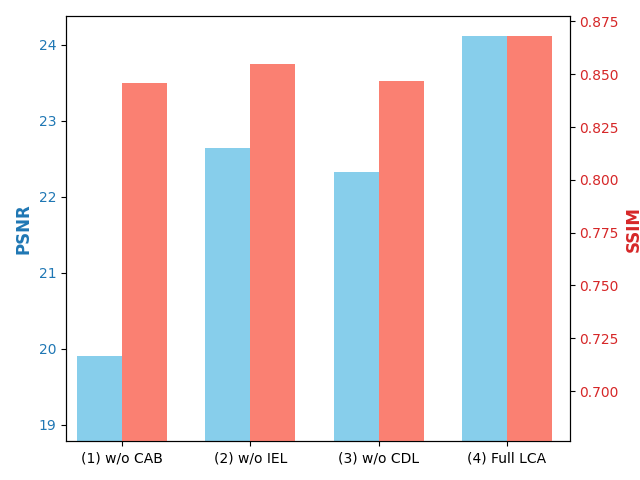}
    \caption{Ablation experiment of the submodules CAB, IEL, and
CDL in LCA. The impact of CAB is significantly higher than the
other two modules, indicating that our designed cross-attention can
better enhance the image to the appropriate brightness level. (Full LCA denotes the origin design of the LCA block.)}
    \label{fig:ab-LCA}
\end{figure}

\subsection{Ablation Study}
We conduct extensive ablation studies to validate our HVI color space and several modules. The evaluations are all performed on the LOLv2-Real dataset for fast convergence and stable performance.

% 去字母，合并同类BCDE
\textbf{Color Spaces.} We conduct an ablation on sRGB, HSV and our HVI on CIDNet as Table \ref{tab:ablation-a}. The ranking of image restoration quality with CIDNet is as follows: HVI, sRGB, and HSV. Comparing row (2), (3), and (6), the best is HVI color space, which exists a significant performance enhancement.

\textbf{Structures and Modules.} First, we examine our LCA module in the present color spaces. As shown in Table \ref{tab:ablation-a} row (1), (2), and (6), LCA with baseline gains 2.088, 5.428 dB on PSNR for sRGB and HVI respectively. Second, in our experimental investigation, we have observed distinct statistical patterns between the I-branch and HV-branch. As shown in Table \ref{tab:ablation-a} row (6) and (7), dividing the Enhancement Network into dual branches enhances 1.159 dB in PSNR. Third, by decoupling the image through HVI, a certain correlation between the values of the I-map and the noise of the HV-map can be discovered. To establish the relationship mapping, we incorporate Cross-Attention into the internal of LCA. As Table \ref{tab:ablation-a} row (8), PSNR and SSIM are both greater than row (7) by 0.952 dB and 0.014. Last, as Fig. \ref{fig:ab-LCA}, removing the CAB, CDL or IEL clearly shows a decrease effect of PSNR and SSIM, which demonstrates the effectiveness of sub-modules in LCA block. The visual quality comparisons are shown in Fig. \ref{fi3}. Specifically, removing CAB leads to unstable brightness enhancement, resulting in local overexposure and artifacts. On the other hand, removing IEL or CDL results in excessively dark brightness, thereby affecting the details.

\begin{figure}
    \centering
    \includegraphics[width=1\linewidth]{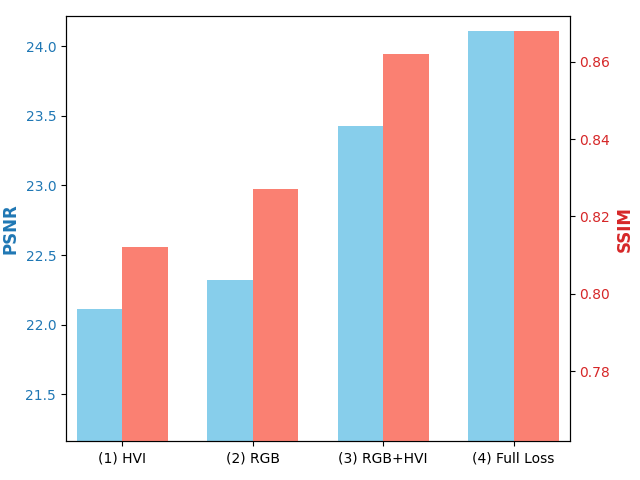}
    \caption{Ablation of losses in our HVI color space, the sRGB color space, and the potential feature space as the perceptual (Perc.) loss with VGG19 network \cite{VGG19}. In the group (4), we incorporate the perceptual loss into both the HVI and sRGB color spaces simultaneously.}
    \label{fig:ab-loss}
\end{figure}

\begin{figure}
    \centering
    \includegraphics[width=1\linewidth]{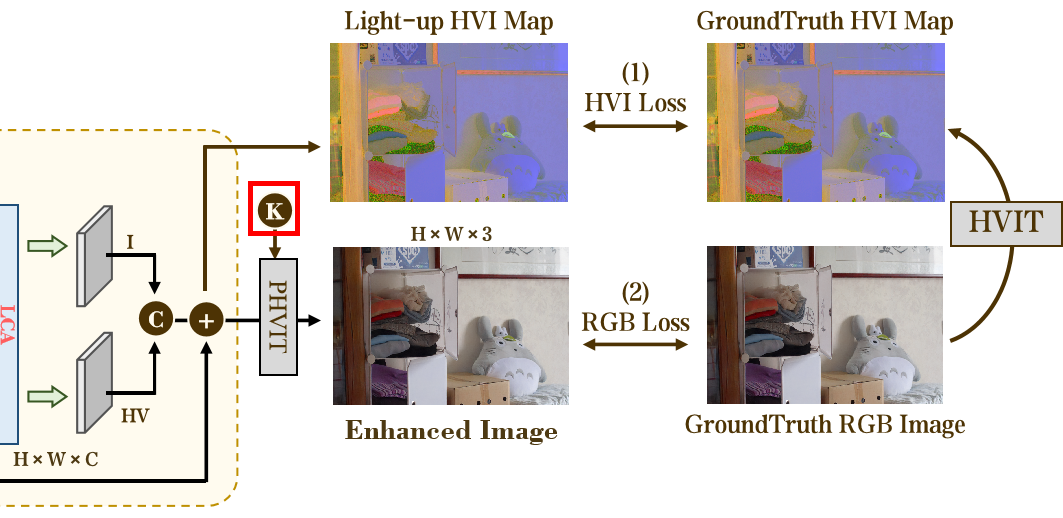}
    \caption{Comparison of loss function computation positions. The trainable parameter $k$ is highlighted within the red box. \textbf{(Zoom in for the best view.)}}
    \label{fig:discussion}
\end{figure}

\textbf{Ablation on loss function.} We conduct an ablation by progressively adding loss on (1)HVI-map, (2)sRGB-image, (3)HVI and sRGB, (4)Perceptual loss on both color space as Table \ref{fig:ab-loss}. Compared to the previous groups, the results of including the loss in all spaces in group (4) shows an improvement in PSNR of 1.998 dB, 1.792 dB, and 0.684 dB. This demonstrates the effectiveness of the loss function we designed.

\textbf{Analysis.} Above experiments sequentially verify the superiority of our the HVI color space, the LCA block, and the dual-branch Enhancement Network with Cross-Attention. For loss ablation experiment, the HVI color space needs to be supervised by the sRGB space losses in order to perform.
% We leave deeper explorations to future works.

\textbf{Loss Discussion: Why is HVI loss performance weaker than RGB loss?} In Table \ref{fig:ab-loss}, the results in group (2) outperform those in the first group. The reason is that the loss functions of HVI and RGB operate at different positions in our network. As presented in Fig. \ref{fig:discussion}, in between the losses of HVI and RGB lies the inverse transformation of HVI, which incorporates a trainable parameter $k$. Using only the HVI loss does not allow $k$ to converge to the optimal solution and the loss in RGB color space is needed to assist in the training process. Therefore, the combination of HVI and RGB losses has achieved better performance as shown in the third group of Fig. \ref{fig:ab-loss}.

% \subsection{Inference Time}
% \begin{table}
%     \centering
%     \renewcommand{\arraystretch}{1.2}
%     \caption{Runtime testing experiments. We input a random tensor with $256\times256$ in three channels and test inference time in above methods. The best result is in\textcolor{red}{~red} color.}
%     \resizebox{\linewidth}{!}{
%     \begin{tabular}{cccccc}
%     \Xhline{1.5pt}
%          \cellcolor{gray!10}Methods& \cellcolor{gray!10}\makecell{Restormer\\\cite{Restormer}}&	 \cellcolor{gray!10}\makecell{SNR-Aware\\ \cite{MIMO}}&	 \cellcolor{gray!10}\makecell{RetinexFormer\\\cite{RetinexFormer}}&	 \cellcolor{gray!10}\makecell{LEDNet\\\cite{LEDNet}}&	 \cellcolor{gray!10}\textbf{CIDNet}\\
%     \Xhline{1.5pt}
         
% \\
%         FLOPs/G$\downarrow$&	131.31& 	26.35 &	15.85& 	35.92& 	\color{red}{7.57} 
% \\
%     \Xhline{1.5pt}
%     \end{tabular}
%     }
    
%     \label{tab:time}
% \end{table}
% Table \ref{tab:time} provides the inference time of different methods. We randomly generated a 256x256 three-channel tensor. The experiments were conducted using a NVIDIA 2080Ti GPU and an AMD Ryzen 9 5900HX with Radeon Graphics CPU. We have a well-performance on CPU time and FLOPs, however, the GPU time consume a little more time comparing to RetinexFormer. The reason is HVIT is not optimized well in Pytorch \cite{paszke2019pytorch} for GPUs. Nevertheless, our CIDNet remains highly competitive and demonstrates superior enhancement results.

\begin{table}[bp]
    \centering
    \renewcommand{\arraystretch}{1.2}
    \caption{Quantitative evaluation on LOL-Blur dataset. PSNR$\uparrow$ and SSIM$\uparrow$: the higher, the better; LPIPS$\downarrow$ and FLOPs$\downarrow$: the lower, the better. The symbol ‘$\dag$’ indicates that we use DeblurGAN-v2 trained on RealBlur \protect\cite{RealBlur} dataset. ‘$\ddag$’ indicates the network is retrained on the LOL-Blur dataset. The highest result is in\textcolor{red}{~red} color.}
    \resizebox{\linewidth}{!}{
\begin{tabular}{l|ccc|c}
\Xhline{1.5pt}
\rowcolor{gray!10} Methods& PSNR$\uparrow$&SSIM$\uparrow$&LPIPS$\downarrow$ &FLOPs$\downarrow$ \\
\Xhline{1.5pt}
ZeroDCE \cite{Zero-DCE} $\rightarrow$ MIMO \cite{MIMO} & 17.680 & 0.542 & 0.422 & - \\

DeblurGAN\textsuperscript{$\dag$} \cite{DeblurGANv2} $\rightarrow$ ZeroDCE \cite{Zero-DCE} & 18.330 & 0.589 & 0.384 & - \\
RetinexFormer\textsuperscript{$\ddag$} \cite{RetinexFormer} & 22.904 & 0.824 & 0.236 & 15.85 \\
MIMO\textsuperscript{$\ddag$} \cite{MIMO} & 24.410 & 0.835 & 0.183 & 62.36 \\
LEDNet\textsuperscript{$\ddag$} \cite{LEDNet} & 25.271 & 0.859 & 0.141 & 35.93 \\
\Xhline{1.5pt}
\textbf{CIDNet}\textsuperscript{$\ddag$} &  \color{red}{26.572} & \color{red}{0.890} & \color{red}{0.120} & \color{red}{7.57} \\
\Xhline{1.5pt}
\end{tabular}
}
    
    \label{tab:blur}
\end{table}
\begin{table}
    \centering
    \renewcommand{\arraystretch}{1.2}
    \caption{Robustness testing experiments. All methods is trained on the LOLv1 and tested in the LOLv2-Syn dataset. The best result is in\textcolor{red}{~red} color.}
    \resizebox{\linewidth}{!}{
    \begin{tabular}{ccccccc}
    \Xhline{1.5pt}
         \cellcolor{gray!10}Methods& \cellcolor{gray!10}LLFlow&\cellcolor{gray!10}RUAS&	 \cellcolor{gray!10}PairLIE&	 \cellcolor{gray!10}RetinexFormer&	 \cellcolor{gray!10}LEDNet&	 \cellcolor{gray!10}\textbf{CIDNet}\\
    \Xhline{1.5pt}
         PSNR$\uparrow$&	17.1191& 	15.3257&\color{red}{19.0743}& 	16.1834& 	16.6210& 	18.6382 
\\
         SSIM$\uparrow$&	0.8117& 	0.4883&0.7965& 	0.7693& 	0.7733& 	\color{red}{0.8200} 
\\
        LPIPS$\downarrow$&	0.2239& 	0.4577&0.2300 &	0.2515& 	0.2196& 	\color{red}{0.2154} 
\\
    \Xhline{1.5pt}
    \end{tabular}
    }
    
    \label{tab:v1-on-v2}
\end{table}

\subsection{Robustness Experiments}
\textbf{LOL-Blur.} Long exposures in dimly lit environments can result in photos that are prone to blurring. To verify the robustness ability of our model, we conduct experiments on the low-light blur dataset LOL-Blur.

In the first set, we perform lighting-up with ZeroDCE and then deblurring with MIMO \cite{MIMO}. In the second set, we perform deblurring with DeblurGAN-v2 \cite{DeblurGANv2} and then light up with ZeroDCE. In the third group, we have retrained on LOL-Blur with four methods, RetinexFormer, MIMO, and LEDNet, and compared them with our CIDNet. The results (as Table \ref{tab:blur}) show that the quantitative comparison of CIDNet against the current stage SOTA method LEDNet by 5.15\%, 3.61\%, and 14.89\% in PSNR, SSIM, and LPIPS metrics respectively. Not only that, the FLOPs of our model are the lowest among these methods.

\begin{figure}
    \centering
    \includegraphics[width=1\linewidth]{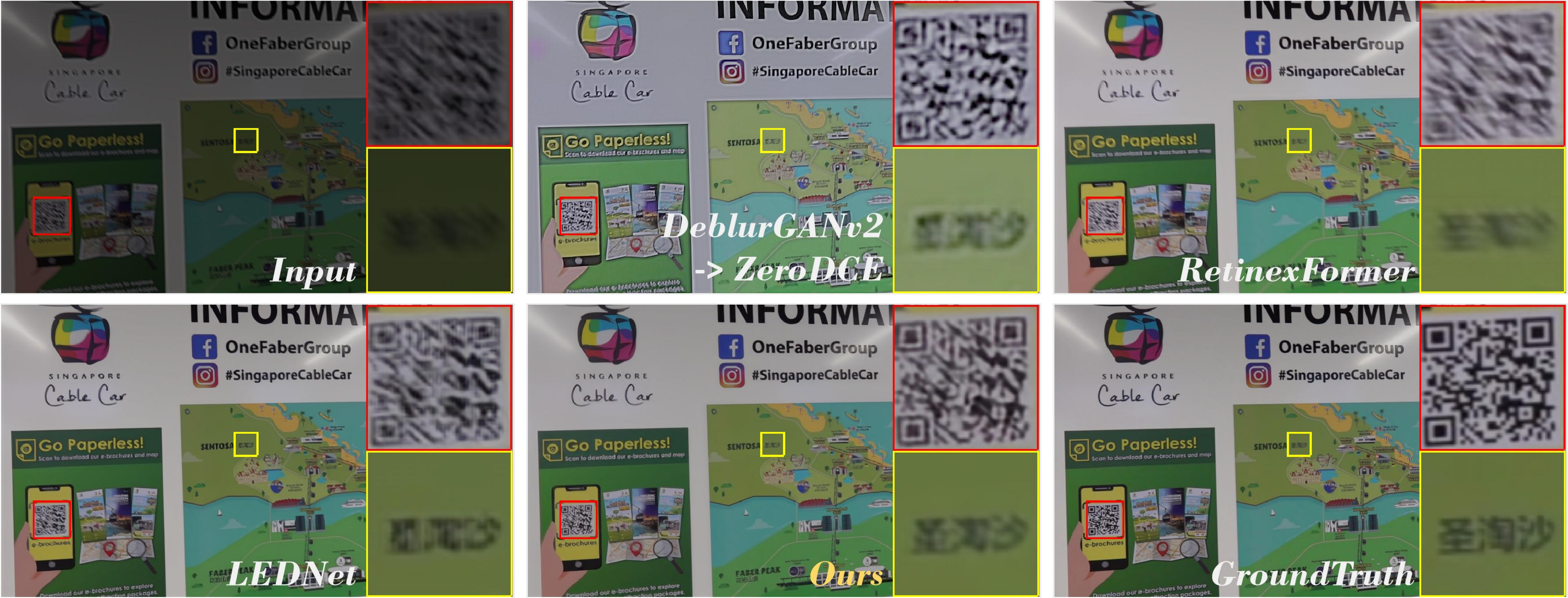}
    \caption{Visual comparison on LOL-Blur dataset. Compared to other methods, our CIDNet is closer to GroundTruth and more dominant in visual recognition.  (\textbf{Zoom in for best view.}) }
    \label{fig:blur}
\end{figure}

As shown in Fig. \ref{fig:blur}, we have taken a set of blurred images, recovered them using different methods, and compared them with GroundTruth. The experimental results reveal that the image reconstructions achieved by CIDNet exhibit a notable improvement in visual comfort and perceptual recognition, thereby enhancing the overall quality and interpretability of the generated images.

\textbf{Cross dataset validation.} To verify the generalization ability of our model, we take the model trained on LOLv1 \cite{RetinexNet} and tested it on LOLv2-Syn \cite{LOLv2} as Table \ref{tab:v1-on-v2}. Compared with three supervised learning SOTA models, LLFlow \cite{LLFlow}, RetinexFormer \cite{RetinexFormer}, and LEDNet \cite{LEDNet}, our model comprehensively outperforms in three metrics. And compared with the unsupervised models PairLIE \cite{PairLIE} and RUAS \cite{RUAS}, which have better generalization ability, our CIDNet also completely outperforms SSIM and LPIPS.\par

\textbf{The effect of HVIT in other methods.} We employ the HVI transform as a versatile module to examine the resilience of our HVI space across multiple models. To this end, the HVIT and PHVIT are incorporated into four supervised models, RetinexFormer, LEDNet, Restormer, and SNR-Aware. Subsequently, these models are retrained and evaluated on the LOLv2-Real \cite{LOLv2} dataset. The outcomes, as illustrated in Table \ref{tab:HVI}, demonstrate the remarkable enhancement achieved by our HVI space in the performance of each method. Impressively, all eleven metrics associated with the four methods exhibit substantial improvements. Moreover, the proposed CIDNet demonstrates superior performance not only in terms of PSNR and SSIM, but also exhibits the shortest CPU (AMD Ryzen 9 5900HX) and GPU (NVIDIA 2080Ti) inference time and minimal computational (FLOPs) overhead.

% \begin{figure*}
%     \centering
%     \includegraphics[width=1\linewidth]{img/types.pdf}
%     \caption{The four convolution schemes I2I, HVI2I, HV2HV, and HVI2HV. Furthermore, it constitude the three HVIT schemes: (1) Half-HVIT: contains I2I and HVI2HV. (2) Separate-HVIT: contains I2I and HV2HV. (3) Full-HVIT: contains HVI2I and VI2HV. (The gray block represent the $3\times3$ convolution layer.}
%     \label{fig:HVIT}
% \end{figure*}

\subsection{Variant HVIT Experiments and Discussion}
To investigate how the first two $3\times3$ convolution layers (in Enhancement Network) learned to generate I-feature and HV-feature, we further develop three different HVIT by changing the inputs. 
As shown in Table \ref{tab:ablation-c} row (1), the Half-HVIT utilized in our pipeline generates intensity features (hereinafter abbreviated as I-Features) from the intensity Map and HV-Features from the HVI-Map. 
We have respectively modified the input of I-Features and HV-Features, leading to the creation of two new HVIT models in Table \ref{tab:ablation-c} row (2) and row (3). The Separate-HVIT replaces the input of HV-Feature with HV-Map, while the Full-HVIT substitutes the input of I-Feature with HVI-Map.

As a result, the Half-HVIT achieves the best restoration results among the three HVIT models. 
The performance drop of the Separate-HVIT is more pronounced, with reductions of 0.377 dB in PSNR, 0.011 in SSIM, and 0.033 in LPIPS. 
This can be attributed to the lower information content in the HV features compared to the HVI-Map, which lacks guidance from the intensity part. On the other hand, the performance decline of the Full-HVIT was due to the interference noise information in the HVI-Map for extracting I-Features, leading to convolutional layers failing to accurately extract key features.

\begin{table}
    \centering
    \renewcommand{\arraystretch}{1.2}
    \caption{HVI transform robustness validation experiments. Embedding HVI transform into other methods and training/testing on LOLv2-Real. Value in brackets represents the amount of better numerical changes with\textcolor{red}{~red} color. The best result of PSNR/SSIM$\uparrow$ and LPIPS/FLOPs$\downarrow$ is in \textbf{bolded}. The lowest time of inference with $256\times256$ in three channels is also in \textbf{bolded}.}
    \resizebox{\linewidth}{!}{
    \begin{tabular}{cccccc}
    \Xhline{1.5pt}
         \cellcolor{gray!10}Methods& \cellcolor{gray!10}RetinexFormer&	 \cellcolor{gray!10}LEDNet&	 \cellcolor{gray!10}Restormer&	 \cellcolor{gray!10}SNR-Aware&	 \cellcolor{gray!10}\textbf{CIDNet}\\
    \Xhline{1.5pt}
         PSNR$\uparrow$&23.600(\textcolor{red}{+0.806})&	23.394(\textcolor{red}{+3.456})&	23.234(\textcolor{red}{+4.541})&	22.251(\textcolor{red}{+0.771})&	\textbf{24.111}
\\
         SSIM$\uparrow$&0.865(\textcolor{red}{+0.025})&	0.837(\textcolor{red}{+0.010})&	0.866(\textcolor{red}{+0.032})&	0.840(-0.009)&	\textbf{0.868}

\\
        LPIPS$\downarrow$&0.113(\textcolor{red}{-0.058})&	0.117(\textcolor{red}{-0.003})&	\textbf{0.093}(\textcolor{red}{-0.022})&	0.117(\textcolor{red}{-0.040})&	0.108
\\
        FLOPs/G$\downarrow$&	15.85&	35.92&	114.25&	26.35&	\textbf{7.57}
\\
        GPU Time/s$\downarrow$&	 	0.062& 0.070& 	0.183& 0.054& 	\textbf{0.053}
\\
         CPU Time/s$\downarrow$& 	0.594& 	0.846& 5.898& 	 1.751& 		\textbf{0.416}
\\
    \Xhline{1.5pt}
    \end{tabular}
    }
    
    \label{tab:HVI}
\end{table}
\begin{table}
    \centering
    \renewcommand{\arraystretch}{1.2}
    \caption{Ablation of three different types of inputs in Enhancement Network. Each distinct convolution layers will extract and generate corresponding intensity features (as I-Feature) and HV-features from different input maps (the type of input is indicated in the columns I-Feature and HV-Feature). The best result is in \textcolor{red}{red} color.}
    \resizebox{\linewidth}{!}{
    \begin{tabular}{l|cc|ccc}
    \Xhline{1.5pt}
         \cellcolor{gray!10}Types&\cellcolor{gray!10}I-Features&\cellcolor{gray!10}HV-Features&\cellcolor{gray!10}PSNR$\uparrow$&	\cellcolor{gray!10}SSIM$\uparrow$&	\cellcolor{gray!10}LPIPS$\downarrow$
\\
\Xhline{1.5pt}
         (1) Half-HVIT& intensity & HVI-Map&\textcolor{red}{24.111}& 	\textcolor{red}{0.868}& 	\textcolor{red}{0.108} 
\\
         (2) Separate-HVIT& intensity & HV-Map&23.734& 	0.857& 	0.141
\\
         (3) Full-HVIT& HVI-Map & HVI-Map&23.814& 	0.859& 	0.127 
\\

\Xhline{1.5pt}
    \end{tabular}
    }
    \label{tab:ablation-c}
\end{table}

\section{Conclusion}
In this paper, we present a novel method for low-light image enhancement using the proposed HVI color space with trainable parameters and the CIDNet to decouple image brightness and color and adapt to various illumination scales. The dual-branch network, build upon the HVI color space, simultaneously processes brightness and color, aided by the plug-and-play LCA module and symmetric HVI Transform module. Our CIDNet outperforms all types of SOTA methods across 11 datasets with lower FLOPs and parameters.

\section*{Acknowledgment}
This work is supported by NSFC of China 62301432, 6230624, Natural Science Basic Research Program of Shaanxi No. 2023-JC-QN-0685, QCYRCXM-2023-057, the Fundamental Research Funds for the Central Universities No. D5000220444.

\par

\bibliographystyle{IEEEtran}
\bibliography{trans}

\vspace{11pt}

% \bf{If you include a photo:}\vspace{-33pt}
% \begin{IEEEbiography}[{\includegraphics[width=1in,height=1.25in,clip,keepaspectratio]{fig1}}]{Michael Shell}
% Use $\backslash${\tt{begin\{IEEEbiography\}}} and then for the 1st argument use $\backslash${\tt{includegraphics}} to declare and link the author photo.
% Use the author name as the 3rd argument followed by the biography text.
% \end{IEEEbiography}

\vspace{11pt}

% \bf{If you will not include a photo:}\vspace{-33pt}
% \begin{IEEEbiographynophoto}{John Doe}
% Use $\backslash${\tt{begin\{IEEEbiographynophoto\}}} and the author name as the argument followed by the biography text.
% \end{IEEEbiographynophoto}

\vfill

\end{document}